\pdfoutput=1

\documentclass[11pt]{article}

\usepackage{ACL2023}

\usepackage{times}
\usepackage{latexsym}

\usepackage[T1]{fontenc}

\usepackage[utf8]{inputenc}

\usepackage{microtype}

\usepackage{inconsolata}

\usepackage{my_commands}
\title{Graph Language Models}

\author{
    Moritz Plenz\\
    Computational Linguistics \\
    Heidelberg University\\
    {\tt \scalebox{1.0}[1.0]{plenz@cl.uni-heidelberg.de}}
    \And
    Anette Frank\\
    Computational Linguistics \\
    Heidelberg University\\
    {\tt \scalebox{1.0}[1.0]{frank@cl.uni-heidelberg.de}}
}

\date{}

\begin{document}
\maketitle
\begin{abstract}

While Language Models (LMs) are the workhorses of NLP, their interplay with structured knowledge graphs (KGs) is still actively researched. Current methods for encoding such graphs typically either (i)~linearize them for embedding with LMs~--~which underutilize structural information, or (ii)~use Graph Neural Networks (GNNs) to preserve the graph structure~--~but GNNs cannot represent text features as well as pretrained LMs. In our work we introduce a novel LM type, the \textit{Graph Language Model} (GLM), that integrates the strengths of both approaches and mitigates their weaknesses. The GLM parameters are initialized from a pretrained LM to enhance understanding of individual graph concepts and triplets. Simultaneously, we design the GLM's architecture to incorporate graph biases, thereby promoting effective knowledge distribution within the graph. This enables GLMs to process graphs, texts, and interleaved inputs of both. Empirical evaluations on relation classification tasks show that GLM embeddings surpass both LM- and GNN-based baselines in supervised and zero-shot setting, demonstrating their versatility.\footnote{\url{https://github.com/Heidelberg-NLP/GraphLanguageModels}}

\end{abstract}

\section{Introduction}

Knowledge Graphs (KGs) are essential for organizing vast data, to facilitate information retrieval, or revealing hidden insights for decision-making~\citep{plenz-etal-2024-pakt}. KGs excel in explicitly representing manifold relationships, so with an expanding wealth of information they become crucial tools in the digital age.

\begin{figure}
    \centering
     \includegraphics[width=0.64\linewidth]{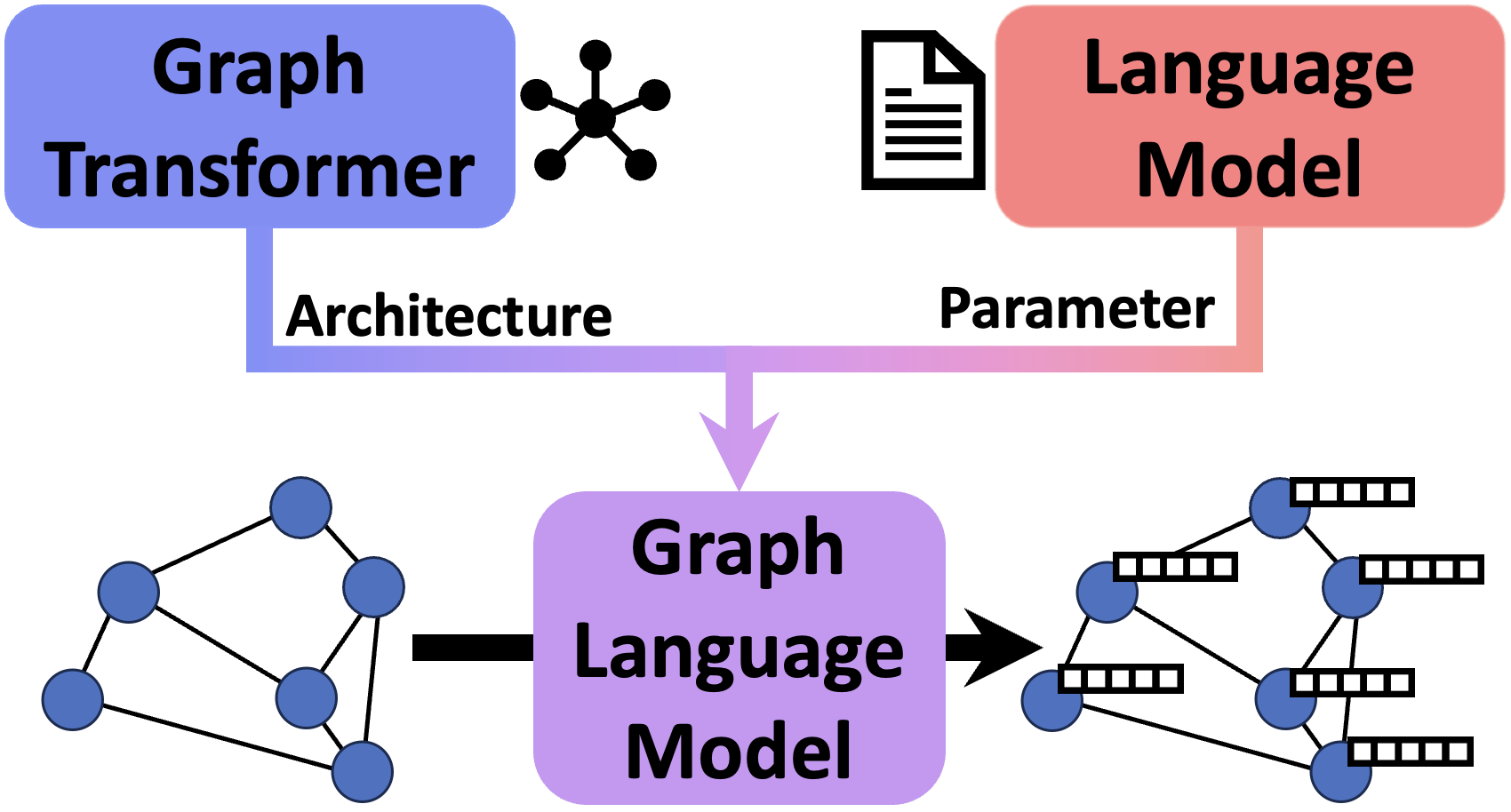}
    \caption{The GLM inherits its architecture from a Graph Transformer, and its parameters from a LM. This enables it to jointly reason over graphs and language.}
    \label{fig:GLM_overview}
\end{figure}

Many KGs consist of knowledge triplets, where nodes are entities and edges represent relationships holding between them. Each triplet represents a fact in pseudo-natural language, e.g., \textit{(Thailand; Capital; Bangkok)} in DBpedia~\citep{auer-etal-2007-dbpedia}. Despite the (usual) simplicity of each individual triplet, complex structures emerge in KGs. We refer to such KGs as Graphs of Triplets (GoTs). 

To use GoTs effectively, we need meaningful encodings of their components. A natural choice is leveraging LMs, as they can capture the semantics of textually encoded entities, relations or entire triplets. But LMs are not prepared to capture graph-structured information and cannot model complex interactions in a GoT. To alleviate this problem, one can leverage graph NNs (GNNs). But GNNs are not well suited to capture meanings associated with text, and hence often LMs are used to convert nodes (and possibly edges) to language-based semantic embeddings. But in such settings, semantic encoding leveraged from LMs and structural reasoning performed by GNNs are separated and are driven by distinct underlying principles. We expect this to limit model performance if both textual and structural information are important for a task. 

In this work we introduce a \textit{Graph Language Model} (GLM) that resolves this tension through early fusion of textual and structural information. Most LMs today are transformers. 
Since transformers operate on sets, \textit{Positional Encoding} (PE) is used to inform them about the inherent sequential ordering of linguistic inputs. 
In our GLM formulation, we modify PE and self-attention to convert LMs (i.e., sequence transformers) to graph transformers that natively operate on graphs, while preserving their LM capabilities. Usually, a new architecture requires pretraining from scratch, which is extremely costly. By adopting some non-invasive changes in the LM's self-attention module, we transform the LM to a Graph Transformer (GT) -- while maintaining compatibility with its pretrained LM parameters. When encoding a graph, LM-like attention patterns process linearly organized textual information of individual triplets, while GT-like attention patterns aggregate information along the graph structure. Hence, the GLM inherits text understanding of triplets from the LM, while its GT architecture allows it to directly perform structural reasoning, without additional GNN layers. 

Importantly, for text sequences -- which can be seen as a special type of graph -- the GLM is identical to the original LM. This allows the GLM to process interleaved inputs of text and GoT jointly, handling both modalities in a single framework. 

Our main contributions are: (i) We propose \textit{Graph Language Models} (GLMs) and a theoretical frame\-work to construct them. GLMs are graph transformers, which enables graph reasoning. Simultaneously, they inherit and exploit LM weights, enabling them to represent and contextualize triplets in a GoT. 
Further, by encoding texts and graph components alike, it can naturally take graph and text data as interleaved inputs. 
(ii) Experiments on relation classification in ConceptNet subgraphs show that GLMs outperform LM- and GNN-based methods for encoding GoTs -- even when the inherited LM parameters are not updated during GLM training.
(iii) KG population experiments on Wikidata subgraphs and corresponding Wikipedia abstracts show that GLMs can reason over interleaved inputs of GoTs and text -- again, outperforming strong LM-based methods.

\section{Related Work}

\paragraph{LMs} 
One way to augment LMs with knowledge from KGs~\citep{pan-etal-2024-unifying} is to formulate pretraining objectives that operate on a KG. E.g., LMs can be trained to generate parts of a KG, encouraging the LM to store KG content in its parameters. Typically, single triplets are used for pretraining~\citep{bosselut-etal-2019-comet,wang-etal-2021-k,hwang-etal-2021-cometatomic2O,west-etal-2023-novacomet}. In such cases, the graph structure is not a target of pretraining. Some works generate larger substructures, such as paths or linearized subgraphs~\citep{wang-etal-2020-connecting,schmitt-etal-2020-unsupervised,huguet-cabot-navigli-2021-rebel}. In either case, the LM needs to memorize the KG, as it will not be part of the input during inference. 

Another approach is to provide the linearized KG as part of the input during inference. This is common for KG-to-text generation~\citep{schmitt-etal-2020-unsupervised,ribeiro-etal-2021-investigating,li-etal-2021-few}, where models learn to take the linearized (typically small-sized) graphs as input. A recent trend is retrieval augmented generation, where relevant parts of a knowledge base, or KG, are retrieved, linearized and provided as part of a prompt~\citep{gao-etal-2024-retrievalaugmented}.\footnote{Cf. \url{www.llamaindex.ai} and \url{www.langchain.com}} 

In both options the graph must be linearized to fit the input or output of a sequence-to-sequence LM. Hence, no graph priors can be enforced -- instead, the LM has to learn the graph structure implicitly. By contrast, GLMs model a graph as a true graph and have inductive graph priors instilled in their architecture. This prepares a GLM for more proficient graph reasoning, compared to a LM approach. 

\paragraph{GNNs}

LMs excel at representing single triplets, but struggle with structural reasoning. To alleviate this problem, LMs can be combined with GNNs. 
Many approaches get node and edge features from LMs and aggregate this information in the graph with GNNs~\citep{lin-etal-2019-kagnet,malaviya-etal-2020-commonsense,zhao-etal-2023-learning}. %
\citet{zhang-etal-2022-greaselm,yasunaga-2022-etal-deep} train models consisting of a LM and a GNN that encode interleaved text and graph inputs jointly. They also use a LM to obtain node features. %

While some approaches jointly train for textual understanding and graph reasoning, none offer a unified method. By contrast, our GLM formulation seamlessly integrates 
both in a holistic framework for embedding language and KGs.

\paragraph{Graph Transformers}

GTs, a special type of GNN~\citep{bronstein-etal-2021-geometric}, gain popularity 
in NLP and beyond~\citep{min-etal-2022-transformer,mueller-etal-2023-attending}. E.g., \citet{koncel-kedziorski-etal-2019-text} and \citet{wang-etal-2020-amr} train GTs to generate text from KGs and AMRs, respectively. Most relevant to our work is \citet{schmitt-etal-2021-modeling}, who use GTs for KG-to-text generation. Similar to us, they employ PE matrices,
but train their model from scratch, which limits its applicability: 
while their model trained on WebNLG~\citep{gardent-etal-2017-webnlg} has a vocabulary size of %
2,100, initializing a GLM from T5, equips it with T5's full vocabulary of 32,128 tokens. 

Concurrently, and independently from our work, \citet{li-etal-2024-unifying} also convert a LM to a graph transformer. 
They focus on data-to-text generation, where they unify table, key-value and KG structures in a unified graph format, and apply structure-enhanced pre-training to support data-to-text generation with their structure-enhanced transformer model. They apply attention maps similar to ours to better capture the graph-structured input, which the pre-trained model rewrites into natural language. %
Contrary to their work, we do not resort to structure-enhanced pre-training -- which is restricted in resources -- but instead assess the GLMs' innate capabilities. %
We showcase the versatility of 
the inherited LM parameters in conjunction with %
our graph transformer architecture, by applying them to challenging reasoning tasks, where the model needs to reason over \textit{complementary} inputs from text and graphs, and where it needs to infer information \textit{not} present in the input, unlike data-to-text generation. %
Moreover, we demonstrate that our architectural changes are highly compatible with the original LM weights, via linear probing experiments, 
where the GLM outperforms conventional LM and Graph Transformer models.

\section{Preliminary: Graph Transformers (GT)} \label{sec:graph_transformer}
This section briefly introduces graph transformers, focusing on architectural choices relevant for our work. We also discuss some general properties of GNNs that motivate our design choices in~\S\ref{sec:glm}. 

The attention in self-attention can be written as 
\begin{equation}
\textrm{softmax}\left(\frac{QK^T}{\sqrt{d}} + B_P + M\right)V,
\end{equation}
where $Q$, $K$ and $V$ are the query, key and value matrices, and $d$ is the query and key dimension. The $B_P$ and $M$ matrices can be used for positional encoding and masking. 
Setting $B_P = M = 0$ yields the standard formulation~\citep{vaswani-etal-2017-attention}. %

\paragraph{Positional Encoding}
The self-attention mechanism of transformer models is permutation invariant, i.e., it doesn't have any notion of the order of its input elements. Thus, positional Encoding (PE) is used  to inform LMs of the ordering of tokens in a text~\citep{dufter-etal-2022-position}. Most approaches employ either \textit{absolute PE}, where absolute token positions are encoded~\citep{vaswani-etal-2017-attention,gehring-etal-2017-convolutional} or \textit{relative PE}, which encodes the relative position between pairs of tokens~\citep{shaw-etal-2018-self,raffel-etal-2020-exploring,su-etal-2021-roformer,press-etal-2022-train}. Absolute PE is typically combined with the input sequence and hence, the PE does not need to be encoded in self-attention ($B_P=0$). For relative PE, $B_P$ encodes a bias depending on the relative distances between pairs of tokens -- for example, by learning one scalar for each possible distance: 
\begin{equation}
    B_P = f(P), 
\end{equation}
where $P$ is a matrix of relative distances and $f(\cdot)$ an elementwise function. %

Similarly, GTs use PEs to encode the structure of the input, and hence, their PE has to encode a graph structure, as opposed to a sequence. This can again be done with absolute or relative PEs. However, defining an ``absolute position'' of a node or edge in a graph is not straightforward. While many methods exist, they are not directly compatible with the usual (absolute) ``counting position'' known from sequence encoding in LMs. In this work we thus focus on relative PE. Given a directed acyclic path in a graph, we can define the (signed) distance between any pair of nodes along a path simply as the number of hops between the nodes. The sign can be set by the direction of the path. Thus, by finding a consistent set of such paths in~\S\ref{sec:glm}, we obtain relative distances and hence the graph's PE. %

\paragraph{Masked Attention} 
In a vanilla transformer, self-attention is computed for all possible pairs of tokens in the input. By contrast, nodes typically only attend to adjacent nodes in GNNs. 
Therefore, information between more distant nodes has to be propagated across multiple GNN layers. For graphs, such sparse message passing approaches are sometimes preferred, as in most graphs the neighborhood size increases exponentially with increasing radius, which can cause loss of information due to over-smoothing~\citep{chen-etal-2020-measuring}. 
Thus, in GTs it can be beneficial to introduce graph priors, for example by restricting self-attention to local neighborhoods. 
This can be realized by setting elements of $M$ to $0$ for pairs of tokens that should be connected, and to $-\infty$ otherwise. 

On the other hand, it has been shown that a global view of the graph can enable efficient, long-ranged information flow~\citep{alon-yahav-2021-on,ribeiro-etal-2020-modeling}. 
We will therefore present two model variants in \S\ref{sec:glm} -- a \textit{local} and a \textit{global} GLM. 

\begin{figure*}  %
    \centering
    \begin{subfigure}[b]{0.35\textwidth}
         \centering
         \includegraphics[width=\textwidth]{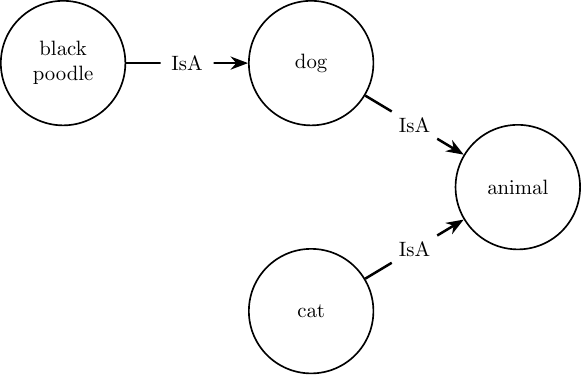}
         \caption{Original GoT.}
         \label{fig:preprocessing_original}
     \end{subfigure}
     \hfill
     \begin{subfigure}[b]{0.6\textwidth}
         \centering
         \includegraphics[width=\textwidth]{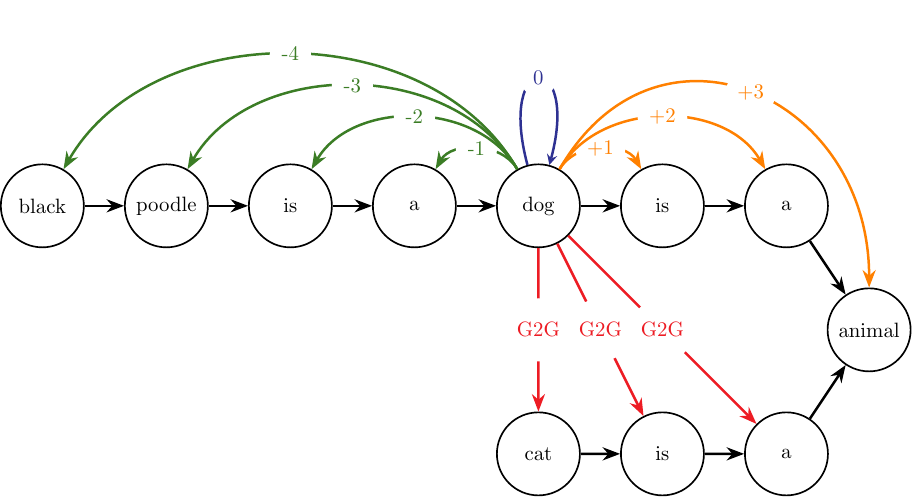} %
         \caption{Extended Levi graph of GoT (with relative distances $P$ for \textit{dog}).}
         \label{fig:preprocessing_levi}
     \end{subfigure}
    
    \caption{Example of graph preprocessing in our GLM. Fig~\ref{fig:preprocessing_levi} shows relative distances for \textit{dog}, i.e., when \textit{dog} is attending to other tokens. The red Graph-to-Graph (G2G) connections only exist for the \gglm, not for the \lglm.}
    \label{fig:preprocessing}
\end{figure*}

\section{Graph Language Model} \label{sec:glm}

\paragraph{GLM vs.\ GT} 
We aim to design an architecture that can efficiently and jointly reason over text and graph-structured data. 
GTs can offer desired graph priors, but they lack language understanding. %

One intuitive approach to bridge this gap is to pretrain a GT from scratch~\citep{schmitt-etal-2021-modeling}. But pretraining is costly and the necessary data bound to be scarce. We thus take a different avenue. We hypothesize that for reasoning over GoTs a model needs language understanding capabilities similar to those used for reasoning over text. Intuitively this should be the case, since (i) GoTs are designed to be understandable by humans and (ii) literate people can ``read'' and understand GoTs. %

By initializing a GT with parameters from a compatible LM, we obtain our Graph Language Model (GLM). The GT architecture introduces graph priors, while parameter initialization from the LM gives it language understanding capabilities. In the following we explain the necessary modifications to the input graph and to the model, to make this work. The general idea is that (verbalized) triplets should resemble natural language as much as possible to enable LM weights to capture them, while graph reasoning should work via message passing. 

\paragraph{Graph preprocessing} 

A LM tokenizer converts text into a sequence of tokens from the LM vocabulary. %
Similarly, we process GoTs, such that the GLM can process the graphs ``as a LM would do'' (cf.\ Fig.~\ref{fig:preprocessing}). To achieve this, we first convert the GoT to its so-called Levi graph~\citep{schmitt-etal-2021-modeling}, i.e., we replace each edge with a node that contains the relation name as text feature, and connect the new node to the head and tail of the original edge via unlabeled edges, preserving the direction of the original edge. %
Next, we tokenize each node %
and split each node into multiple nodes, such that every new node corresponds to a single token. New edges connect adjacent nodes, again preserving the original direction. This yields the extended Levi graph (see Fig.~\ref{fig:preprocessing_levi}). In this representation, each triplet is represented as a sequence of tokens -- just as it would be for a standard LM.\footnote{Note that the token sequence of the converted GoT is not necessarily perfectly identical to the token sequence that corresponds to the input triplets. We tokenize each node in the Levi graph individually, to ensure consistent tokenization of concepts shared by multiple triplets. This removes whitespace between concepts and edges, which impacts tokenization. We leave  investigation of the impact of this effect to future work.}

\begin{figure}
    \centering
     \includegraphics[width=0.8\linewidth]{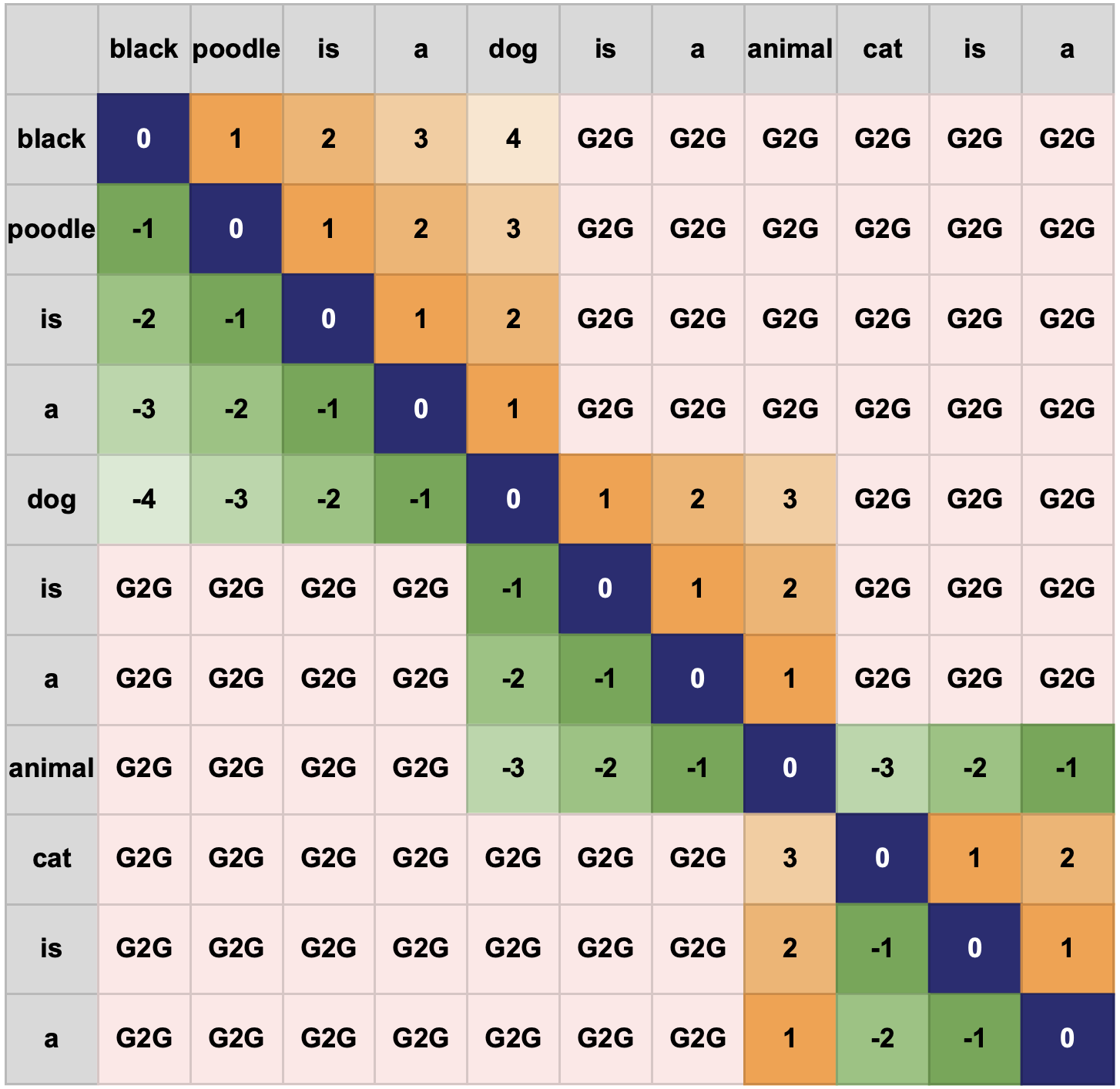}
    \caption{Relative position matrix $P$ for tokens in Fig.~\ref{fig:preprocessing_levi}. Entries with \textit{G2G} have no relative position (\lglm) or are initialized from $+\infty$ (\gglm). Cf. \S\ref{sec:app:model}. %
    }
    \label{fig:PE_matrix}
\end{figure}

\paragraph{Positional Encodings} 
As discussed in~\S\ref{sec:graph_transformer}, we prefer PEs that encode the relative position between pairs of tokens, determined by their signed distance. We can directly adopt this method to encode the relative position between pairs of tokens occurring within the same triplet -- by simply considering the triplet as a piece of text, and counting the token distance in this text. Note that a single token can occur in multiple triplets, leading to, e.g., multiple ``left-hand side neighbors'' (cf. \textit{animal} in Fig.~\ref{fig:preprocessing_levi} and~\ref{fig:PE_matrix}). While this does not occur in ordinary sequential text, it does not impose a problem for relative PE. 

Yet, the approach above breaks when tokens do not belong to the same triplet. To determine the distance between such pairs of tokens, previous work considered, e.g., the length of the shortest path between them~\citep{schmitt-etal-2021-modeling}. However, this results in PEs that do not come natural to a LM, since a triplet would appear in reversed order, if it is traversed in ``the wrong direction'' in the shortest path.\footnote{For example, \textit{cat} would see the graph as the following sequence: \textit{cat is a animal a is dog a is poodle black}.} 
We therefore omit structure-informed PE between tokens that do not belong to the same triplet and instead propose two GLM variants: a local (\lglm) and a global (\gglm) one.

\paragraph{Local and global GLM}
Fig.~\ref{fig:PE_matrix} shows the relative token position matrix $P$ for the graph in Fig.~\ref{fig:preprocessing_levi}. In the \lglm\ the self-attention mechanism is restricted to tokens from the same triplet. This means that attention to any token located beyond the local triplet is set to $0$ -- and hence does not require PE. Still, in such configurations, messages can propagate through the graph across multiple layers, since tokens belonging to a concept can be shared by multiple triplets. This is analogous to standard message passing in GNNs, where non-adjacent nodes have no direct connection, but can still share information via message passing. For example, the representation of \textit{dog} is contextualized by the triplets \textit{black poodle is a dog} and \textit{dog is a animal} after the first \lglm\ layer. Hence in the second layer, when \textit{animal} attends to \textit{dog}, the \textit{animal} embedding gets impacted by \textit{black poodle}, even though there is no direct connection from \textit{animal} to \textit{black poodle}. %

However, it has been shown that a global view can have benefits~\citep{ribeiro-etal-2020-modeling}. Hence, we also formalize the \gglm, as an alternative where self-attention can connect any node to every other node. For this setting we need to assign a PE to any pair of tokens, including those that do not occur within the same triplet. For these pairs we introduce a new graph-to-graph (G2G) relative position. LMs don't have learned parameters for G2G connections, so we initialize the parameters with the corresponding parameters of a relative position of $+\infty$. In a LM a relative position of $+\infty$ means that the respective tokens occur somewhere ``far'' away in a remote text passage. LMs learn a \textit{proximity bias} during pretraining, i.e., they tend to have higher attention scores between tokens that are close to each other in the text. This means that tokens with a high relative distance tend to have low attention scores. For our \gglm\ this corresponds to a graph bias where distant nodes are less important, but are still accessible.\footnote{Preliminary experiments showed that initializing G2G parameters from $+\infty$ outperforms random initialization, which outperforms initialization from $0$.} Note that unlike in the \lglm, this bias is \textit{not} part of the architecture. It originates from the pretrained parameters, meaning that the \gglm\ can learn to attend to distant tokens.

Along with $P$ and $M$, the GLM takes a sequence of all tokens in the extended Levi graph as input. For this, we technically need to ``linearize'' the graph. However, the order of tokens in the resulting sequence does not matter: relative positions in $P$ are determined by distances in the graph, not in the sequence. Permuting the input sequence simply means that rows and columns of $P$ and $M$ need to be permuted accordingly, but the resulting token embeddings remain unchanged. See example matrices for $P$ and $M$ for \lglm\ and \gglm\ in \S\ref{sec:app:model}. 

Being transformers, GLMs have the same computational complexity as their respective LM. For sparse graphs the \lglm\ could make use of sparse matrix multiplication, making it more efficient than a corresponding LM or \gglm. However, for our experiments this was not necessary. 

\paragraph{Joint graph and text encoding} If we use normal matrices for $P$ and $M$, the GLM is identical to its underlying LM. Hence, GLMs can be applied to texts and -- more interestingly -- interleaved inputs of text and graph. In this joint setting, $P$ and $M$ each consists of four sub-matrices that correspond to self-attention between tokens from (i) graph-to-graph, (ii) text-to-text, (iii) text-to-graph and (iv) graph-to-text. Graph-to-graph sub-matrices are formatted as described above for \lglm\ and \gglm, respectively. Text-to-text sub-matrices are standard matrices from conventional sequence transformers. We introduce new T2G and G2T relative positions for text-to-graph, and graph-to-text connections, respectively. With this, the model can learn interaction strength between the two modalities. Similar to G2G in \gglm, we initialize T2G and G2T parameters from $+\infty$. See example matrices in \S\ref{sec:app:model}. 

\paragraph{Uni- and Bidirectional LMs}
If a LM's self-atten\-tion is unidirectional, information can only propagate along the direction of arrows in Fig.~\ref{fig:preprocessing_levi} for the \lglm. This means that, e.g., the representation of the node \textit{black poodle} is independent of the rest of the graph. We could augment the graph with inverse relations to enable bidirectional information flow with unidirectional LMs, but in this work, we restrict our analysis to bidirectional models. 

\paragraph{T5}
We use T5~\citep{raffel-etal-2020-exploring} -- a bidirectional encoder with unidirectional decoder -- as base LM to instantiate GLMs. In T5, relative distances in $P$ group into so-called buckets, and each bucket maps to one learned positional bias in $B_P$ for each head. Positional biases are shared across layers. The decoder is not needed to encode graphs, but 
can be used to generate sequences, such as 
text or linearized graphs in future work. 

\section{Experiments}

We assess the GLMs' capabilities for embedding GoTs in two experiments on relation (label) classification, i.e., classifying which relation belongs to a given head and tail entity. One experiment uses ConceptNet~\citep[CN;][]{speer-etal-2017-conceptnet} subgraphs that we construct to enable analysis of the impact of structural graph properties. In a second experiment on Wikidata~\citep{vrandecic-etal-2014-wikidata} subgraphs and associated Wikipedia abstracts we test GLMs on interleaved inputs of text and graph.

\subsection{Representing and reasoning over Graphs}
\label{sec:exp:cn}
We construct a balanced dataset of
English CN subgraphs consisting 
of 13,600 train, 1,700 dev and 1,700 test instances with 17 distinct relations as labels. We replace the relation to be predicted with \texttt{<extra\_id\_0>}, T5's first mask token. 

To investigate the impact of varying graph complexities, we experiment with different graph sizes denoted by their radius $r$. We ensure that small graphs are strict subgraphs of larger graphs, such that potential performance gains in larger graphs must stem from additional long-ranged context. 

To evaluate model effectiveness when long-ranged connections are crucial, we mask complete subgraphs around the relation to be predicted. 
The size of a masked subgraph is $m$, where $m=0$ means no mask, $m=1$ masks neighboring concepts, $m=2$ masks neighboring concepts and the next relations, etc. We replace each masked concept and relation with a different mask token. Construction details and statistics are shown in \S\ref{sec:app:exp:cn:dataset}. 

\subsubsection{Experimental setup} \label{sec:exp:cn:setup}

The input to our model is a CN subgraph. The relation to be predicted is replaced with \texttt{<extra\_id\_0>}. The GLM encodes the graphs as in \S\ref{sec:glm}, producing an embedding for each token. 
A linear classification head gives the final prediction from the mask's embedding. 
We verbalize unmasked relations using static templates~\citep{plenz-etal-2023-similarity}, shown in~\S\ref{sec:app:experiments}, Table \ref{tab:app:verbalization_templates_cn}. 

In a \textbf{finetuning} setting we train the GLM and the classification head jointly. However, since the GLM is initialized from a LM, we hypothesize that it should produce meaningful embeddings, even without any training. To test this hypothesis, we train only the classification head, i.e., we only train a \textbf{linear probe}. %
In this setting, the GLM was never trained on any graph data, similar to a zero-shot setting. The linear probe only extracts linear features and hence, can only achieve high performance if the GLM embeddings show expressive features. 

We report mean accuracy across 5 different runs. See \S\ref{sec:app:exp:cn:setup} for hyperparameters. Unless stated otherwise, we use T5-small to allow many baselines. %

\begin{table*} %
    \centering
    \resizebox{\linewidth}{!}{
    \begin{tabular}{ccccccccccccc}
    \toprule
        & \multirow{2}{*}{Model} 
        & $r$ & 1 & 2 & 3 & 4 & 5 & 4 & 4 & 4 & 4 & 4 \\
        & & $m$ & 0 & 0 & 0 & 0 & 0 & 1 & 2 & 3 & 4 & 5 \\ \cmidrule(){1-1} \cmidrule(lr){2-3} \cmidrule(lr){4-8} \cmidrule(l){9-13}

        \parbox[t]{2mm}{\multirow{4}{*}{\rotatebox[origin=c]{90}{Lin. Prob.}}}
        & \multicolumn{2}{c}{\lglm} & 
        \textbf{55.4$\pm$0.3} & 57.1$\pm$0.3 & 56.8$\pm$0.6 & 56.9$\pm$0.4 & 57.0$\pm$0.4 & 30.4$\pm$0.4 & 17.8$\pm$0.2 & 14.0$\pm$0.3 & 11.4$\pm$0.5 & 11.9$\pm$0.3 \\
        & \multicolumn{2}{c}{\gglm} & 
        \textbf{55.4$\pm$0.3} & \textbf{58.6$\pm$0.7} & \textbf{58.8$\pm$0.6} & \textbf{59.3$\pm$0.7} & \textbf{59.5$\pm$0.4} & \textbf{41.8$\pm$0.8} & \textbf{25.6$\pm$0.9} & \textbf{22.0$\pm$0.6} & \textbf{19.4$\pm$0.5} & \textbf{17.0$\pm$0.2} \\
        & \multicolumn{2}{c}{T5 (list)} & 
        53.7$\pm$0.3 & 56.8$\pm$1.1 & 56.5$\pm$1.2 & 55.8$\pm$0.6 & 55.3$\pm$0.5 & 20.3$\pm$0.6 & 19.9$\pm$0.4 & 15.3$\pm$0.6 & 14.0$\pm$1.1 & 10.2$\pm$1.2 \\
        & \multicolumn{2}{c}{T5 (set)} & 
        53.1$\pm$0.6 & 52.8$\pm$1.2 & 54.6$\pm$0.6 & 53.9$\pm$0.5 & 53.1$\pm$0.8 & 18.2$\pm$0.6 & 16.7$\pm$0.5 & 13.1$\pm$0.7 & 12.3$\pm$0.6 & 9.7$\pm$0.9 \\

        \cmidrule(){1-1} \cmidrule(lr){2-3} \cmidrule(lr){4-8} \cmidrule(l){9-13}
        
        \parbox[t]{2mm}{\multirow{8}{*}{\rotatebox[origin=c]{90}{Finetuning}}}
        & \multicolumn{2}{c}{\lglm} & 
        64.0$\pm$1.3 & 64.0$\pm$1.0 & 64.4$\pm$0.7 &\textbf{ 64.1$\pm$0.9} & 64.2$\pm$1.1 & 47.9$\pm$0.4 & 26.8$\pm$0.8 & 23.8$\pm$0.9 & 19.8$\pm$1.1 & 18.1$\pm$0.7 \\
        & \multicolumn{2}{c}{\gglm} & 
        63.2$\pm$0.9 & 64.4$\pm$1.1 & 64.6$\pm$1.2 & \textbf{64.1$\pm$1.3} & \textbf{65.3$\pm$0.7} & \textbf{48.0$\pm$0.6} & \textbf{27.2$\pm$0.7} & \textbf{24.2$\pm$0.7} & \textbf{20.2$\pm$1.4} & \textbf{19.2$\pm$0.7} \\
        & \multicolumn{2}{c}{T5 (list)} & 
        \textbf{64.9$\pm$1.0} & 64.9$\pm$1.2 & \textbf{64.9$\pm$1.3} & 63.9$\pm$0.9 & 64.0$\pm$0.6 & 40.4$\pm$0.8 & 21.8$\pm$0.8 & 17.8$\pm$1.0 & 15.4$\pm$0.3 & 12.8$\pm$0.5 \\
        & \multicolumn{2}{c}{T5 (set)} & 
        63.9$\pm$0.7 & \textbf{65.8$\pm$0.8} & 64.0$\pm$0.3 & \textbf{64.1$\pm$1.2} & 64.3$\pm$1.1 & 40.3$\pm$1.2 & 21.8$\pm$0.7 & 18.0$\pm$0.6 & 15.5$\pm$0.6 & 13.1$\pm$0.7 \\
        & \multicolumn{2}{c}{GCN} & 
        44.3$\pm$0.9 & 37.1$\pm$1.0 & 34.4$\pm$1.2 & 36.5$\pm$0.6 & 36.8$\pm$1.4 & 22.2$\pm$1.2 & 21.9$\pm$0.8 & 12.1$\pm$3.5 & 9.0$\pm$4.3 & 5.9$\pm$0.0 \\
        & \multicolumn{2}{c}{GAT} & 
        44.5$\pm$0.9 & 40.6$\pm$1.3 & 36.3$\pm$1.3 & 37.0$\pm$0.8 & 37.0$\pm$0.8 & 20.0$\pm$0.7 & 20.8$\pm$0.2 & 14.0$\pm$0.6 & 13.8$\pm$0.8 & 11.0$\pm$0.6 \\
        & \multicolumn{2}{c}{\lgt} & 
        24.2$\pm$3.4 & 35.0$\pm$1.2 & 34.7$\pm$1.3 & 32.7$\pm$2.9 & 34.5$\pm$2.8 & 30.1$\pm$2.6 & 12.8$\pm$2.4 & 15.5$\pm$0.3 & 9.5$\pm$1.3 & 10.0$\pm$1.6 \\
        & \multicolumn{2}{c}{\ggt} & 
        27.6$\pm$1.9 & 29.0$\pm$0.8 & 23.4$\pm$1.2 & 19.2$\pm$1.2 & 15.6$\pm$1.5 & 18.6$\pm$0.7 & 13.2$\pm$1.1 & 14.5$\pm$0.6 & 12.4$\pm$1.3 & 12.1$\pm$1.7 \\
    \bottomrule
    \end{tabular}
    }
    \caption{Relation label classification accuracy on CN in \%. Results are shown for \textit{Linear Probing} and \textit{Finetuning}.}
    \label{tab:res}
\end{table*}

\subsubsection{Baselines} \label{sec:exp:cn:baselines}
We compare to several baselines inspired by related work. For all baselines we utilize the T5 encoder as underlying LM. This allows us to focus on the architectural design of different model types.

\paragraph{LM} 
For LM-based approaches we linearize the input graphs to a sequence, by concatenating the verbalized triplets. 
There are structured ways to linearize graphs, but such graph traversals generally require the graph to be directed and acyclic -- which makes them inapplicable to linearizing GoTs. Instead, we order the triplets either randomly (\textbf{T5 set}) or alphabetically (\textbf{T5 list}). For \textit{T5 set}, triplets are shuffled randomly in every training epoch such that the model can learn to generalize to unseen orderings. The concatenated triplets are passed to the T5 encoder, and the embedding of \texttt{<extra\_id\_0>} is presented to the classification head. 

\paragraph{GNN} For GNN baselines we encode each node of the original graph (cf. Fig.~\ref{fig:preprocessing_original}) with the T5 encoder, and train a GNN using these static embeddings. After the final layer, the GNN returns 17 logits for each node. As final logits, we take the mean logit of the two nodes adjacent to the relation to be predicted. 
We experiment with different variants as GNN layers: \textbf{GCN}~\citep{kipf-welling-2017-semi} and \textbf{GAT}~\citep{velickovic-etal-2018-graph}. %
Since GNNs do not come with pretrained weights, we only apply them in finetuning, when training all parameters.

\paragraph{Graph transformer} Finally we compare GLMs to models with the same architecture, but random weight initialization (normal graph transformers). This allows us to assess the impact of weight initialization from a LM with two further baselines: $\bm{\ell}$\textbf{GT} and $\bm{g}$\textbf{GT}. 
We only consider GTs with finetuning.

\subsubsection{Results}
\paragraph{Linear probing}
Tab.~\ref{tab:res} shows the relation label prediction accuracy for linear probing, i.e., when training only the classification head. 
Our first observation is that \gglm\ is consistently the best, outperforming \lglm\ and the LM baselines. For a radius of $r=1$ we have exactly one triplet, which has almost the same representation in the GLM and LM approaches. 
The only difference is that 
the LM baselines have an end-of-sentence token, which the GLM does not have. Surprisingly, not having the end-of-sentence token seems to be an advantage with 
linear probing,
but 
we will see later that this changes when updating model weights. 

For $r\geq3$, LM baselines show decreasing performance with increasing radii. By contrast, both \lglm\ and \gglm\ show increasing performances with increasing radii. This indicates that GLMs can utilize the additional context. %
But LM baselines don't have any inbuilt methods to grasp distances in the graph, which could cause them to fail at distinguishing relevant from less relevant information. 

The performance gap between \gglm\ and LM models tends to increase for larger $m$, i.e., when larger sub-structures are masked. However, the \lglm\ under\-per\-forms for large $m$, highlighting the ad\-van\-tage of the global view in \gglm\ when long-ranged connections are necessary. 

The overall high performance of GLMs confirms our assumption that GLMs are compatible with LM weights, even without any training. Increasing performance with increasing radii further shows that GLMs have good inductive graph biases. When long-range connections are relevant, the representations learned by \gglm\ outperform the locally constrained \lglm\ -- which showcases the strength of the global view that the \gglm\ is able to take. 

\paragraph{Finetuning}
Tab.~\ref{tab:res} shows results when training all parameters. In this setting, models can adjust to the task and learn to reason over graphs through parameter updates. In addition, GLMs can tune parameters to better match the novel input structure. 

The GLM and LM variants are consistently better than GNN and GT methods, which indicates that linguistic understanding is potentially more important than graph reasoning for this task. Models outperform their linear probing scores, which shows that finetuning is, as expected, beneficial. 

Overall, the GLMs perform best, while GTs perform the worst. The only difference between the two model groups is weight initialization -- the GLMs are initialized from T5, while the GTs are randomly initialized. Further, we observe that for $r\geq1$ and $m=0$ the local GT (\lgt) significantly outperforms its global counterpart \ggt. For the GLM the global version is on par, or even better than the local one. This shows the effectiveness of T5's attention mechanism: thanks to its weight initialization, \gglm\ attends to relevant tokens even in large context windows, while \ggt\ suffers from potentially distracting long-ranged information. 

For $m=0$ the differences between GLM and LM approaches are small, with a slight trend for GLMs to outperform LMs on large graphs, and vice versa for small graphs. 
However, when graph reasoning is more important due to masking ($m\geq1$), then GLMs consistently and significantly outperform all other baselines. This indicates that LMs can learn to do simple graph reasoning through parameter updates, but underperform in more complex graph reasoning tasks where either graphs are larger, or long-ranged connections are required. 

For $m\geq1$, the \gglm\ outperforms \lglm\ due to its global connections. In contrast to the linear probing setting, the \lglm\ outperforms other baselines for all non-zero levels of masking. This indicates that \lglm\ can learn to use long-ranged information during training, if the task requires it. %

\paragraph{Impact of model size} To investigate the effect of model size, we train the most promising approaches (GLM and LM) in 3 different sizes. Tab.~\ref{tab:res_rel_allparams_modelsize} in \S\ref{sec:app:exp:cn:results} shows that overall larger models perform better. %
Surprisingly, the base models sometimes outperform the larger models for settings that require more graph reasoning, i.e., larger $m$. However, these differences are small and non-significant. In most cases, \gglm\ large or base are the best model.

\begin{figure}
    \centering
     \begin{subfigure}[b]{0.5\textwidth}
         \centering
         \includegraphics[width=\textwidth]{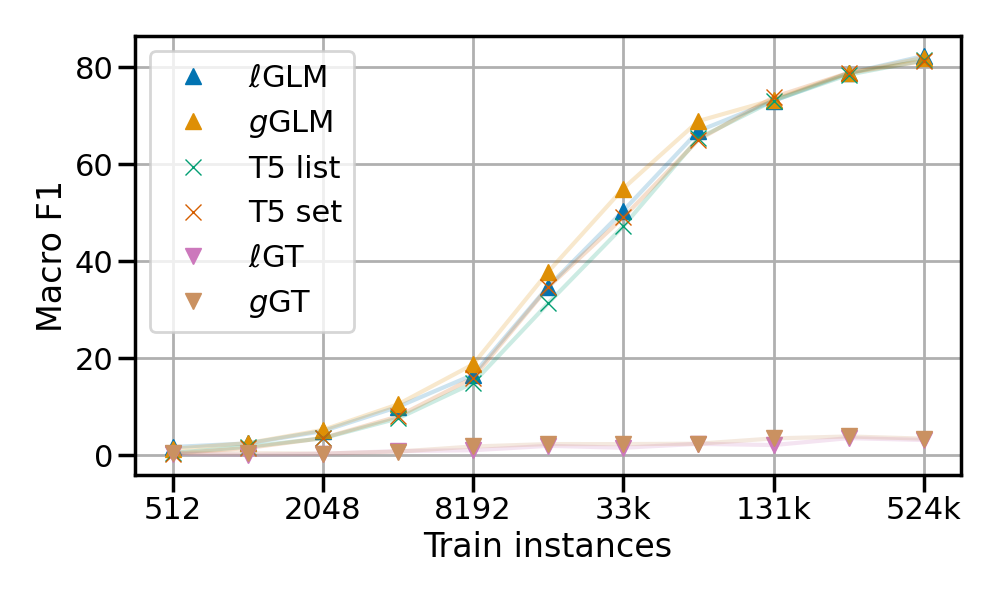}
         \caption{Relation label classification.}
         \label{fig:exp:wiki:baselines:rel_class}
     \end{subfigure}
     \hfill
     \begin{subfigure}[b]{0.5\textwidth}
         \centering
         \includegraphics[width=\textwidth]{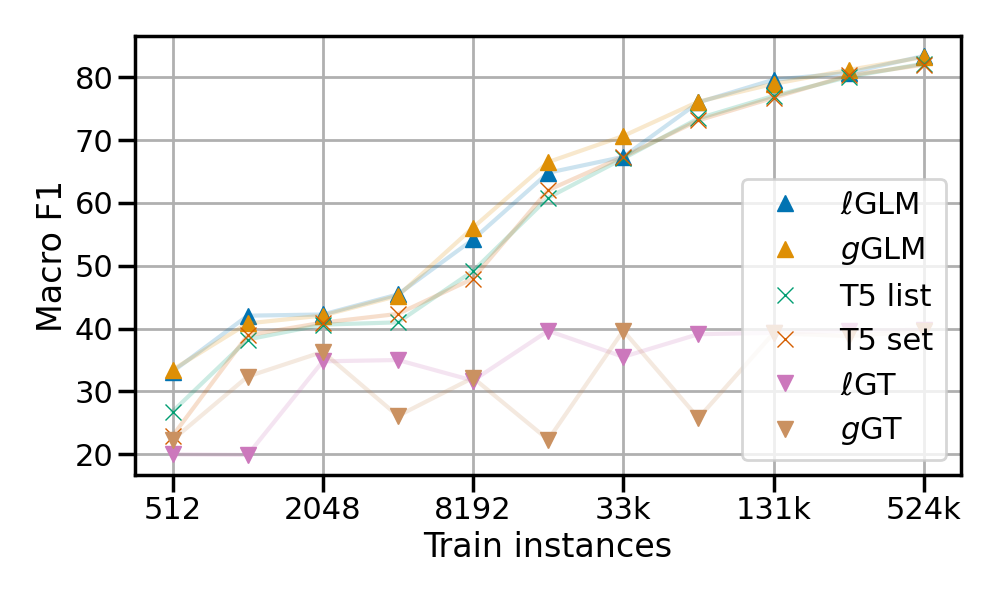}
         \caption{Source classification.}
     \end{subfigure}
    \caption{KG population test results during training. \gglm\ outperforms T5 set by up to 6 points in \ref{fig:exp:wiki:baselines:rel_class}.}
    \label{fig:exp:wiki:baselines}
\end{figure}

\subsection{Jointly representing Graph and Text} 
\label{sec:exp:wiki}

We now investigate 
GLM capabilities to process interleaved inputs of text and graph in a KG population setup, i.e., extending a KG with new relation instances. Subtask 1 performs text-guided \textit{relation} classification where some relations may be inferrable from the text, while others may exploit %
graph knowledge to make predictions. In Subtask 2, models classify the \textit{source} of a predicted relation, i.e., whether it 
can be inferred from the text, 
or whether it requires (additional) graph knowledge.

We construct our data from \citet{huguet-cabot-navigli-2021-rebel}, who offer a corpus of Wikipedia abstracts that are linked to Wikidata via entity linking. Their focus is relation extraction, so they filter the graphs using NLI, such that all triplets are entailed by the text. We augment the entailed triplets with further triplets from Wikidata that are not entailed by the text. 
For a given text, subgraph, head and tail entity, models will jointly predict the \textit{relation} and the \textit{source}. We adopt the 220 most common relations in our train graphs and a ``no-relation'' label. For source labels we have 3 classes: entailed by the text, not entailed and no-relation. No-relation is the correct label iff the relation is also no-relation. 
\S\ref{sec:app:exp:wiki:dataset} shows statistics and construction details. 

\subsubsection{Experimental setup and baselines}

Unlike \S\ref{sec:exp:cn:setup}, models now receive text \textit{and} graph data as input. We train two distinct classification heads on the mask's embedding for relation and source classification. While the mask is part of the graph, its embedding depends on both modalities. 
The final loss is the sum of the relation classification and the source prediction loss, weighted by 0.9 and 0.1. We use T5-large, but otherwise baselines are as in \S\ref{sec:exp:cn:baselines}. \S\ref{sec:app:exp:wiki:setup} shows the training details. 

\subsubsection{Results} 
Fig.~\ref{fig:exp:wiki:baselines} and Tab.~\ref{tab:app:exp:wiki:scores} show test set performance for a) relation and b) source classification, at different training stages. \gglm\ performs the best overall, followed by \lglm. %
LM baselines are competitive, but lag behind at early stages and for source prediction. Again, GT baselines perform poorly, showcasing the advantage of weight initialization in GLM -- even with large-scale training data. For all models, training plateaus beyond $\sim$~500k seen instances (cf.\ Fig.~\ref{fig:app:exp:wiki:long_train_curves} in \S\ref{sec:app:exp:wiki:res}), so we stop training at this cut-off.

Tab.~\ref{tab:exp:wiki:ablation} gives results for ablating different input modalities to GLMs. Since source prediction always requires text input, we test relation classification w/o source prediction. 
Ablating the text or graph lowers performance by similar amounts, indicating that GLMs utilize both modalities. Training curves in Fig.~\ref{fig:app:exp:wiki:ablation} reveal that first, the model almost exclusively utilizes text data, but quickly learns to make use of the graph. For textually entailed triplets, text is more impactful than the graph, and vice versa for other triplets (cf. Tab.~\ref{tab:app:exp:wiki:ablation_text_graph}). Ablating graphs lowers source prediction by $\sim$4.5 points, which shows that GLMs benefit from graph information even for predominantly text oriented tasks.

The results show that GLMs can efficiently reason over interleaved inputs of graph and text, especially with limited training data. This makes GLMs a promising new model type for knowledge-intense NLP tasks, such as KG population or Q\&A.

\begin{table} %
    \centering
    \resizebox{.9\linewidth}{!}{
    \begin{tabular}{lcccc}
    \toprule
        \multirow{2}{*}{Ablation} & \multicolumn{2}{c}{Relation classification} & \multicolumn{2}{c}{Source classification} \\
        & \lglm & \gglm & \lglm & \gglm \\
        \cmidrule(r){1-1} \cmidrule(lr){2-3} \cmidrule(l){4-5}
        w/ text \& graph & 82.63 & 82.25 & 83.39 & 83.21 \\
        w/o text & -6.22 & -5.84 & -- & -- \\
        w/o graph & -6.05 & -5.10 & -4.67 & -4.49 \\
        w/o text \& graph & -19.62\phantom{-} & -19.24\phantom{-} & -- & -- \\
    \bottomrule
    \end{tabular}
    }
    \caption{Ablations for KG population in macro F1. %
    } 
    \label{tab:exp:wiki:ablation}
\end{table}

\section{Conclusion}
We present the Graph Language Model (GLM) -- a graph transformer initialized with weights from a LM. 
It excels at graph reasoning, while simultaneously encoding textual triplets in the graph as LMs do, thereby bridging the gap between LMs and GNNs. %
GLMs can natively reason over joint inputs from texts and graphs, leveraging and enhancing each modality. 
Experiments show the GLM's advantage over LM and GNN based baselines, even in a linear probing setting. %
In particular, GLMs greatly outperform graph transformers. 
This highlights the need for pretrained LM weights, even for graph reasoning. 
We therefore advocate GLMs as a valuable tool for advancing research in embedding and leveraging knowledge graphs for NLP tasks. 

\section*{Limitations}
While GLMs are designed as general purpose tools for knowledge-intense NLP tasks, our evaluation is limited to English knowledge graphs. However, we explore various types of knowledge graphs (commonsense and factual) and tasks (relation classification, text-guided relation classification, and source prediction), broadening our empirical assessment. 
Confirming GLMs improved text and graph reasoning skills for different languages, domains and tasks is left for future work. %

Our GLM framework supports instantiation from any LM with relative positional encoding, including rotary positional encoding. Comprehensive comparisons to determine the most suitable models for the GLM framework remain for future investigation. Nonetheless, bidirectional LMs are expected to perform best in the novel framework, because unidirectional LMs necessitate additional inverse relations, as discussed in \S\ref{sec:glm}.

\section*{Ethical considerations}
We do not foresee immediate ethical concerns for our research, as we rely on well-established datasets. However, even established datasets can contain undesirable biases which our method could potentially spread and amplify. 

Looking ahead, our focus lies in enriching knowledge graph integration within language models, with the aim of enhancing factuality and mitigating hallucination. This advancement is expected to bolster the reliability and controllability of LMs, leading to positive societal impacts. Furthermore, LMs relying on knowledge graphs may facilitate easier maintenance, potentially reducing the need for frequent retraining of deployed models, thereby promoting sustainability in NLP practices.

\section*{Acknowledgements}
We want to thank Letiția Pârcălăbescu for providing feedback on our manuscript.

This work was funded by DFG, the German Research Foundation, within the project ``ACCEPT: Perspectivized Argument Knowledge Graphs for Deliberation'', as part of the priority program ``RATIO: Robust Argumentation Machines'' (SPP-1999).

\bibliography{anthology,custom}

\begin{thebibliography}{41}
\expandafter\ifx\csname natexlab\endcsname\relax\def\natexlab#1{#1}\fi

\bibitem[{Alon and Yahav(2021)}]{alon-yahav-2021-on}
Uri Alon and Eran Yahav. 2021.
\newblock \href {https://openreview.net/forum?id=i80OPhOCVH2} {On the
  bottleneck of graph neural networks and its practical implications}.
\newblock In \emph{International Conference on Learning Representations}.

\bibitem[{Auer et~al.(2007)Auer, Bizer, Kobilarov, Lehmann, Cyganiak, and
  Ives}]{auer-etal-2007-dbpedia}
S{\"o}ren Auer, Christian Bizer, Georgi Kobilarov, Jens Lehmann, Richard
  Cyganiak, and Zachary Ives. 2007.
\newblock Dbpedia: A nucleus for a web of open data.
\newblock In \emph{The Semantic Web}, pages 722--735, Berlin, Heidelberg.
  Springer Berlin Heidelberg.

\bibitem[{Bosselut et~al.(2019)Bosselut, Rashkin, Sap, Malaviya, Celikyilmaz,
  and Choi}]{bosselut-etal-2019-comet}
Antoine Bosselut, Hannah Rashkin, Maarten Sap, Chaitanya Malaviya, Asli
  Celikyilmaz, and Yejin Choi. 2019.
\newblock \href {https://doi.org/10.18653/v1/P19-1470} {{COMET}: Commonsense
  transformers for automatic knowledge graph construction}.
\newblock In \emph{Proceedings of the 57th Annual Meeting of the Association
  for Computational Linguistics}, pages 4762--4779, Florence, Italy.
  Association for Computational Linguistics.

\bibitem[{Bronstein et~al.(2021)Bronstein, Bruna, Cohen, and
  Veli{\v{c}}kovi{\'c}}]{bronstein-etal-2021-geometric}
Michael~M Bronstein, Joan Bruna, Taco Cohen, and Petar Veli{\v{c}}kovi{\'c}.
  2021.
\newblock Geometric deep learning: Grids, groups, graphs, geodesics, and
  gauges.
\newblock \emph{arXiv preprint arXiv:2104.13478}.

\bibitem[{Chen et~al.(2020)Chen, Lin, Li, Li, Zhou, and
  Sun}]{chen-etal-2020-measuring}
Deli Chen, Yankai Lin, Wei Li, Peng Li, Jie Zhou, and Xu~Sun. 2020.
\newblock \href {https://doi.org/10.1609/aaai.v34i04.5747} {Measuring and
  relieving the over-smoothing problem for graph neural networks from the
  topological view}.
\newblock \emph{Proceedings of the AAAI Conference on Artificial Intelligence},
  34(04):3438--3445.

\bibitem[{Dufter et~al.(2022)Dufter, Schmitt, and
  Sch{\"u}tze}]{dufter-etal-2022-position}
Philipp Dufter, Martin Schmitt, and Hinrich Sch{\"u}tze. 2022.
\newblock \href {https://doi.org/10.1162/coli_a_00445} {Position information in
  transformers: An overview}.
\newblock \emph{Computational Linguistics}, 48(3):733--763.

\bibitem[{Gao et~al.(2024)Gao, Xiong, Gao, Jia, Pan, Bi, Dai, Sun, Wang, and
  Wang}]{gao-etal-2024-retrievalaugmented}
Yunfan Gao, Yun Xiong, Xinyu Gao, Kangxiang Jia, Jinliu Pan, Yuxi Bi, Yi~Dai,
  Jiawei Sun, Meng Wang, and Haofen Wang. 2024.
\newblock \href {http://arxiv.org/abs/2312.10997} {Retrieval-augmented
  generation for large language models: A survey}.

\bibitem[{Gardent et~al.(2017)Gardent, Shimorina, Narayan, and
  Perez-Beltrachini}]{gardent-etal-2017-webnlg}
Claire Gardent, Anastasia Shimorina, Shashi Narayan, and Laura
  Perez-Beltrachini. 2017.
\newblock \href {https://doi.org/10.18653/v1/W17-3518} {The {W}eb{NLG}
  challenge: Generating text from {RDF} data}.
\newblock In \emph{Proceedings of the 10th International Conference on Natural
  Language Generation}, pages 124--133, Santiago de Compostela, Spain.
  Association for Computational Linguistics.

\bibitem[{Gehring et~al.(2017)Gehring, Auli, Grangier, and
  Dauphin}]{gehring-etal-2017-convolutional}
Jonas Gehring, Michael Auli, David Grangier, and Yann Dauphin. 2017.
\newblock \href {https://doi.org/10.18653/v1/P17-1012} {A convolutional encoder
  model for neural machine translation}.
\newblock In \emph{Proceedings of the 55th Annual Meeting of the Association
  for Computational Linguistics (Volume 1: Long Papers)}, pages 123--135,
  Vancouver, Canada. Association for Computational Linguistics.

\bibitem[{Huguet~Cabot and Navigli(2021)}]{huguet-cabot-navigli-2021-rebel}
Pere-Llu{\'\i}s Huguet~Cabot and Roberto Navigli. 2021.
\newblock \href {https://doi.org/10.18653/v1/2021.findings-emnlp.204} {{REBEL}:
  Relation extraction by end-to-end language generation}.
\newblock In \emph{Findings of the Association for Computational Linguistics:
  EMNLP 2021}, pages 2370--2381, Punta Cana, Dominican Republic. Association
  for Computational Linguistics.

\bibitem[{Hwang et~al.(2021)Hwang, Bhagavatula, {Le Bras}, Da, Sakaguchi,
  Bosselut, and Choi}]{hwang-etal-2021-cometatomic2O}
Jena~D. Hwang, Chandra Bhagavatula, Ronan {Le Bras}, Jeff Da, Keisuke
  Sakaguchi, Antoine Bosselut, and Yejin Choi. 2021.
\newblock Comet-atomic 2020: On symbolic and neural commonsense knowledge
  graphs.
\newblock In \emph{AAAI}.

\bibitem[{Kipf and Welling(2017)}]{kipf-welling-2017-semi}
Thomas~N. Kipf and Max Welling. 2017.
\newblock Semi-supervised classification with graph convolutional networks.
\newblock In \emph{International Conference on Learning Representations
  (ICLR)}.

\bibitem[{Koncel-Kedziorski et~al.(2019)Koncel-Kedziorski, Bekal, Luan, Lapata,
  and Hajishirzi}]{koncel-kedziorski-etal-2019-text}
Rik Koncel-Kedziorski, Dhanush Bekal, Yi~Luan, Mirella Lapata, and Hannaneh
  Hajishirzi. 2019.
\newblock \href {https://doi.org/10.18653/v1/N19-1238} {{T}ext {G}eneration
  from {K}nowledge {G}raphs with {G}raph {T}ransformers}.
\newblock In \emph{Proceedings of the 2019 Conference of the North {A}merican
  Chapter of the Association for Computational Linguistics: Human Language
  Technologies, Volume 1 (Long and Short Papers)}, pages 2284--2293,
  Minneapolis, Minnesota. Association for Computational Linguistics.

\bibitem[{Li et~al.(2021)Li, Tang, Zhao, Wei, Yuan, and Wen}]{li-etal-2021-few}
Junyi Li, Tianyi Tang, Wayne~Xin Zhao, Zhicheng Wei, Nicholas~Jing Yuan, and
  Ji-Rong Wen. 2021.
\newblock {F}ew-shot {K}nowledge {G}raph-to-{T}ext {G}eneration with
  {P}retrained {L}anguage {M}odels.
\newblock In \emph{ACL Findings}.

\bibitem[{Li et~al.(2024)Li, Li, Geng, Yang, Li, Yuan, He, Yuan, Ma, Huang, and
  Li}]{li-etal-2024-unifying}
Shujie Li, Liang Li, Ruiying Geng, Min Yang, Binhua Li, Guanghu Yuan, Wanwei
  He, Shao Yuan, Can Ma, Fei Huang, and Yongbin Li. 2024.
\newblock \href {https://doi.org/10.1162/tacl_a_00641} {{Unifying Structured
  Data as Graph for Data-to-Text Pre-Training}}.
\newblock \emph{Transactions of the Association for Computational Linguistics},
  12:210--228.

\bibitem[{Lin et~al.(2019)Lin, Chen, Chen, and Ren}]{lin-etal-2019-kagnet}
Bill~Yuchen Lin, Xinyue Chen, Jamin Chen, and Xiang Ren. 2019.
\newblock \href {https://doi.org/10.18653/v1/D19-1282} {{K}ag{N}et:
  Knowledge-aware graph networks for commonsense reasoning}.
\newblock In \emph{Proceedings of the 2019 Conference on Empirical Methods in
  Natural Language Processing and the 9th International Joint Conference on
  Natural Language Processing (EMNLP-IJCNLP)}, pages 2829--2839, Hong Kong,
  China. Association for Computational Linguistics.

\bibitem[{Malaviya et~al.(2020)Malaviya, Bhagavatula, Bosselut, and
  Choi}]{malaviya-etal-2020-commonsense}
Chaitanya Malaviya, Chandra Bhagavatula, Antoine Bosselut, and Yejin Choi.
  2020.
\newblock Commonsense knowledge base completion with structural and semantic
  context.
\newblock \emph{Proceedings of the 34th AAAI Conference on Artificial
  Intelligence}.

\bibitem[{Min et~al.(2022)Min, Chen, Bian, Xu, Zhao, bing Huang, Zhao, Huang,
  Ananiadou, and Rong}]{min-etal-2022-transformer}
Erxue Min, Runfa Chen, Yatao Bian, Tingyang Xu, Kangfei Zhao, Wen bing Huang,
  Peilin Zhao, Junzhou Huang, Sophia Ananiadou, and Yu~Rong. 2022.
\newblock \href {https://api.semanticscholar.org/CorpusID:246904638}
  {Transformer for graphs: An overview from architecture perspective}.
\newblock \emph{ArXiv}, abs/2202.08455.

\bibitem[{Müller et~al.(2023)Müller, Morris, Galkin, and
  Ramp\'{a}\v{s}ek}]{mueller-etal-2023-attending}
Luis Müller, Christopher Morris, Mikhail Galkin, and Ladislav
  Ramp\'{a}\v{s}ek. 2023.
\newblock {Attending to Graph Transformers}.
\newblock \emph{Arxiv preprint}.

\bibitem[{Pan et~al.(2024)Pan, Luo, Wang, Chen, Wang, and
  Wu}]{pan-etal-2024-unifying}
Shirui Pan, Linhao Luo, Yufei Wang, Chen Chen, Jiapu Wang, and Xindong Wu.
  2024.
\newblock Unifying large language models and knowledge graphs: A roadmap.
\newblock \emph{IEEE Transactions on Knowledge and Data Engineering (TKDE)}.

\bibitem[{Plenz et~al.(2024)Plenz, Heinisch, Frank, and
  Cimiano}]{plenz-etal-2024-pakt}
Moritz Plenz, Philipp Heinisch, Anette Frank, and Philipp Cimiano. 2024.
\newblock Pakt: Perspectivized argumentation knowledge graph and tool for
  deliberation analysis.

\bibitem[{Plenz et~al.(2023)Plenz, Opitz, Heinisch, Cimiano, and
  Frank}]{plenz-etal-2023-similarity}
Moritz Plenz, Juri Opitz, Philipp Heinisch, Philipp Cimiano, and Anette Frank.
  2023.
\newblock \href {https://doi.org/10.18653/v1/2023.acl-long.338}
  {Similarity-weighted construction of contextualized commonsense knowledge
  graphs for knowledge-intense argumentation tasks}.
\newblock In \emph{Proceedings of the 61st Annual Meeting of the Association
  for Computational Linguistics (Volume 1: Long Papers)}, pages 6130--6158,
  Toronto, Canada. Association for Computational Linguistics.

\bibitem[{Press et~al.(2022)Press, Smith, and Lewis}]{press-etal-2022-train}
Ofir Press, Noah Smith, and Mike Lewis. 2022.
\newblock \href {https://openreview.net/forum?id=R8sQPpGCv0} {Train short, test
  long: Attention with linear biases enables input length extrapolation}.
\newblock In \emph{International Conference on Learning Representations}.

\bibitem[{Raffel et~al.(2020)Raffel, Shazeer, Roberts, Lee, Narang, Matena,
  Zhou, Li, and Liu}]{raffel-etal-2020-exploring}
Colin Raffel, Noam Shazeer, Adam Roberts, Katherine Lee, Sharan Narang, Michael
  Matena, Yanqi Zhou, Wei Li, and Peter~J. Liu. 2020.
\newblock \href {http://jmlr.org/papers/v21/20-074.html} {Exploring the limits
  of transfer learning with a unified text-to-text transformer}.
\newblock \emph{Journal of Machine Learning Research}, 21(140):1--67.

\bibitem[{Ribeiro et~al.(2021)Ribeiro, Schmitt, Sch{\"u}tze, and
  Gurevych}]{ribeiro-etal-2021-investigating}
Leonardo F.~R. Ribeiro, Martin Schmitt, Hinrich Sch{\"u}tze, and Iryna
  Gurevych. 2021.
\newblock \href {https://doi.org/10.18653/v1/2021.nlp4convai-1.20}
  {Investigating pretrained language models for graph-to-text generation}.
\newblock In \emph{Proceedings of the 3rd Workshop on Natural Language
  Processing for Conversational AI}, pages 211--227, Online. Association for
  Computational Linguistics.

\bibitem[{Ribeiro et~al.(2020)Ribeiro, Zhang, Gardent, and
  Gurevych}]{ribeiro-etal-2020-modeling}
Leonardo F.~R. Ribeiro, Yue Zhang, Claire Gardent, and Iryna Gurevych. 2020.
\newblock \href {https://doi.org/10.1162/tacl_a_00332} {Modeling global and
  local node contexts for text generation from knowledge graphs}.
\newblock \emph{Transactions of the Association for Computational Linguistics},
  8:589--604.

\bibitem[{Schmitt et~al.(2021)Schmitt, Ribeiro, Dufter, Gurevych, and
  Sch{\"u}tze}]{schmitt-etal-2021-modeling}
Martin Schmitt, Leonardo F.~R. Ribeiro, Philipp Dufter, Iryna Gurevych, and
  Hinrich Sch{\"u}tze. 2021.
\newblock \href {https://doi.org/10.18653/v1/2021.textgraphs-1.2} {Modeling
  graph structure via relative position for text generation from knowledge
  graphs}.
\newblock In \emph{Proceedings of the Fifteenth Workshop on Graph-Based Methods
  for Natural Language Processing (TextGraphs-15)}, pages 10--21, Mexico City,
  Mexico. Association for Computational Linguistics.

\bibitem[{Schmitt et~al.(2020)Schmitt, Sharifzadeh, Tresp, and
  Sch{\"u}tze}]{schmitt-etal-2020-unsupervised}
Martin Schmitt, Sahand Sharifzadeh, Volker Tresp, and Hinrich Sch{\"u}tze.
  2020.
\newblock \href {https://doi.org/10.18653/v1/2020.emnlp-main.577} {An
  unsupervised joint system for text generation from knowledge graphs and
  semantic parsing}.
\newblock In \emph{Proceedings of the 2020 Conference on Empirical Methods in
  Natural Language Processing (EMNLP)}, pages 7117--7130, Online. Association
  for Computational Linguistics.

\bibitem[{Shaw et~al.(2018)Shaw, Uszkoreit, and Vaswani}]{shaw-etal-2018-self}
Peter Shaw, Jakob Uszkoreit, and Ashish Vaswani. 2018.
\newblock \href {https://doi.org/10.18653/v1/N18-2074} {Self-attention with
  relative position representations}.
\newblock In \emph{Proceedings of the 2018 Conference of the North {A}merican
  Chapter of the Association for Computational Linguistics: Human Language
  Technologies, Volume 2 (Short Papers)}, pages 464--468, New Orleans,
  Louisiana. Association for Computational Linguistics.

\bibitem[{Speer et~al.(2017)Speer, Chin, and
  Havasi}]{speer-etal-2017-conceptnet}
Robyn Speer, Joshua Chin, and Catherine Havasi. 2017.
\newblock Conceptnet 5.5: An open multilingual graph of general knowledge.
\newblock In \emph{Proceedings of the Thirty-First AAAI Conference on
  Artificial Intelligence}, AAAI'17, page 4444–4451. AAAI Press.

\bibitem[{Su et~al.(2021)Su, Lu, Pan, Wen, and Liu}]{su-etal-2021-roformer}
Jianlin Su, Yu~Lu, Shengfeng Pan, Bo~Wen, and Yunfeng Liu. 2021.
\newblock \href {http://arxiv.org/abs/2104.09864} {Roformer: Enhanced
  transformer with rotary position embedding}.
\newblock \emph{CoRR}, abs/2104.09864.

\bibitem[{Vaswani et~al.(2017)Vaswani, Shazeer, Parmar, Uszkoreit, Jones,
  Gomez, Kaiser, and Polosukhin}]{vaswani-etal-2017-attention}
Ashish Vaswani, Noam Shazeer, Niki Parmar, Jakob Uszkoreit, Llion Jones,
  Aidan~N Gomez, \L~ukasz Kaiser, and Illia Polosukhin. 2017.
\newblock \href
  {https://proceedings.neurips.cc/paper_files/paper/2017/file/3f5ee243547dee91fbd053c1c4a845aa-Paper.pdf}
  {Attention is all you need}.
\newblock In \emph{Advances in Neural Information Processing Systems},
  volume~30. Curran Associates, Inc.

\bibitem[{Veli{\v{c}}kovi{\'{c}} et~al.(2018)Veli{\v{c}}kovi{\'{c}}, Cucurull,
  Casanova, Romero, Li{\`{o}}, and Bengio}]{velickovic-etal-2018-graph}
Petar Veli{\v{c}}kovi{\'{c}}, Guillem Cucurull, Arantxa Casanova, Adriana
  Romero, Pietro Li{\`{o}}, and Yoshua Bengio. 2018.
\newblock \href {https://openreview.net/forum?id=rJXMpikCZ} {{Graph Attention
  Networks}}.
\newblock \emph{International Conference on Learning Representations}.

\bibitem[{Vrande\v{c}i\'{c} and
  Kr\"{o}tzsch(2014)}]{vrandecic-etal-2014-wikidata}
Denny Vrande\v{c}i\'{c} and Markus Kr\"{o}tzsch. 2014.
\newblock \href {https://doi.org/10.1145/2629489} {Wikidata: A free
  collaborative knowledgebase}.
\newblock \emph{Commun. ACM}, 57(10):78–85.

\bibitem[{Wang et~al.(2020{\natexlab{a}})Wang, Peng, Ilievski, Szekely, and
  Ren}]{wang-etal-2020-connecting}
Peifeng Wang, Nanyun Peng, Filip Ilievski, Pedro Szekely, and Xiang Ren.
  2020{\natexlab{a}}.
\newblock \href {https://doi.org/10.18653/v1/2020.findings-emnlp.369}
  {Connecting the dots: A knowledgeable path generator for commonsense question
  answering}.
\newblock In \emph{Findings of the Association for Computational Linguistics:
  EMNLP 2020}, pages 4129--4140, Online. Association for Computational
  Linguistics.

\bibitem[{Wang et~al.(2021)Wang, Tang, Duan, Wei, Huang, Ji, Cao, Jiang, and
  Zhou}]{wang-etal-2021-k}
Ruize Wang, Duyu Tang, Nan Duan, Zhongyu Wei, Xuanjing Huang, Jianshu Ji,
  Guihong Cao, Daxin Jiang, and Ming Zhou. 2021.
\newblock \href {https://doi.org/10.18653/v1/2021.findings-acl.121}
  {{K-Adapter}: {I}nfusing {K}nowledge into {P}re-{T}rained {M}odels with
  {A}dapters}.
\newblock In \emph{Findings of the Association for Computational Linguistics:
  ACL-IJCNLP 2021}, pages 1405--1418, Online. Association for Computational
  Linguistics.

\bibitem[{Wang et~al.(2020{\natexlab{b}})Wang, Wan, and
  Jin}]{wang-etal-2020-amr}
Tianming Wang, Xiaojun Wan, and Hanqi Jin. 2020{\natexlab{b}}.
\newblock \href {https://doi.org/10.1162/tacl_a_00297} {{AMR}-to-text
  generation with graph transformer}.
\newblock \emph{Transactions of the Association for Computational Linguistics},
  8:19--33.

\bibitem[{West et~al.(2023)West, Bras, Sorensen, Lin, Jiang, Lu, Chandu,
  Hessel, Baheti, Bhagavatula, and Choi}]{west-etal-2023-novacomet}
Peter West, Ronan Bras, Taylor Sorensen, Bill Lin, Liwei Jiang, Ximing Lu,
  Khyathi Chandu, Jack Hessel, Ashutosh Baheti, Chandra Bhagavatula, and Yejin
  Choi. 2023.
\newblock \href {https://aclanthology.org/2023.findings-emnlp.80}
  {{N}ova{COMET}: Open commonsense foundation models with symbolic knowledge
  distillation}.
\newblock In \emph{Findings of the Association for Computational Linguistics:
  EMNLP 2023}, pages 1127--1149, Singapore. Association for Computational
  Linguistics.

\bibitem[{Yasunaga et~al.(2022)Yasunaga, Bosselut, Ren, Zhang, Manning, Liang,
  and Leskovec}]{yasunaga-2022-etal-deep}
Michihiro Yasunaga, Antoine Bosselut, Hongyu Ren, Xikun Zhang, Christopher~D
  Manning, Percy Liang, and Jure Leskovec. 2022.
\newblock \href {https://openreview.net/forum?id=4NpoSrT8uU-} {Deep
  bidirectional language-knowledge graph pretraining}.
\newblock In \emph{Advances in Neural Information Processing Systems}.

\bibitem[{Zhang et~al.(2022)Zhang, Bosselut, Yasunaga, Ren, Liang, Manning, and
  Leskovec}]{zhang-etal-2022-greaselm}
Xikun Zhang, Antoine Bosselut, Michihiro Yasunaga, Hongyu Ren, Percy Liang,
  Christopher~D Manning, and Jure Leskovec. 2022.
\newblock \href {https://openreview.net/forum?id=41e9o6cQPj} {Grease{LM}: Graph
  {REAS}oning enhanced language models}.
\newblock In \emph{International Conference on Learning Representations}.

\bibitem[{Zhao et~al.(2023)Zhao, Qu, Li, Yan, Liu, Li, Xie, and
  Tang}]{zhao-etal-2023-learning}
Jianan Zhao, Meng Qu, Chaozhuo Li, Hao Yan, Qian Liu, Rui Li, Xing Xie, and
  Jian Tang. 2023.
\newblock \href {https://openreview.net/forum?id=q0nmYciuuZN} {Learning on
  large-scale text-attributed graphs via variational inference}.
\newblock In \emph{The Eleventh International Conference on Learning
  Representations}.

\end{thebibliography}
\bibliographystyle{acl_natbib}

\appendix

\section{Model} \label{sec:app:model}
\begin{figure*}  %
     \centering
     \begin{subfigure}[b]{0.6\textwidth}
         \centering
         \includegraphics[width=\textwidth]{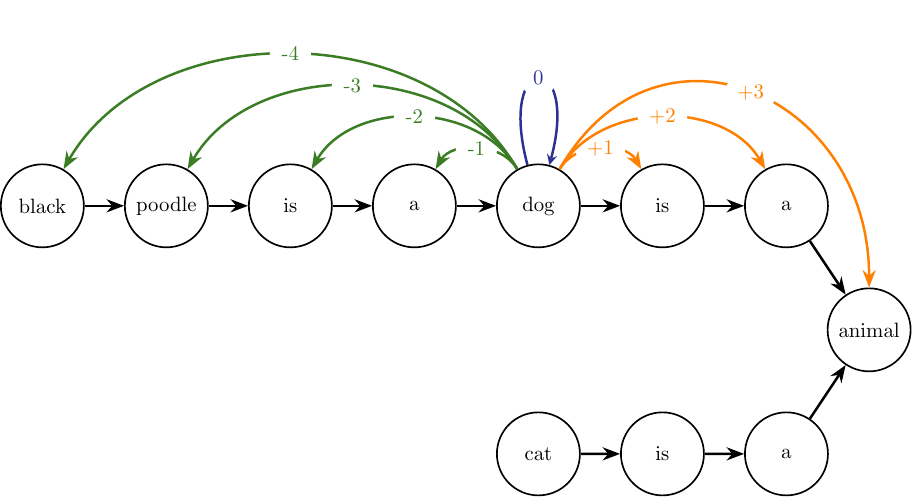} %
         \caption{Relative positions $P$ for \textit{dog} in \lglm.}
         \label{fig:app:lglm_levi}
     \end{subfigure}
    \\ \vspace{0.4cm}
    \centering
    \begin{subfigure}[b]{0.315\textwidth}
         \centering
         \includegraphics[width=\textwidth]{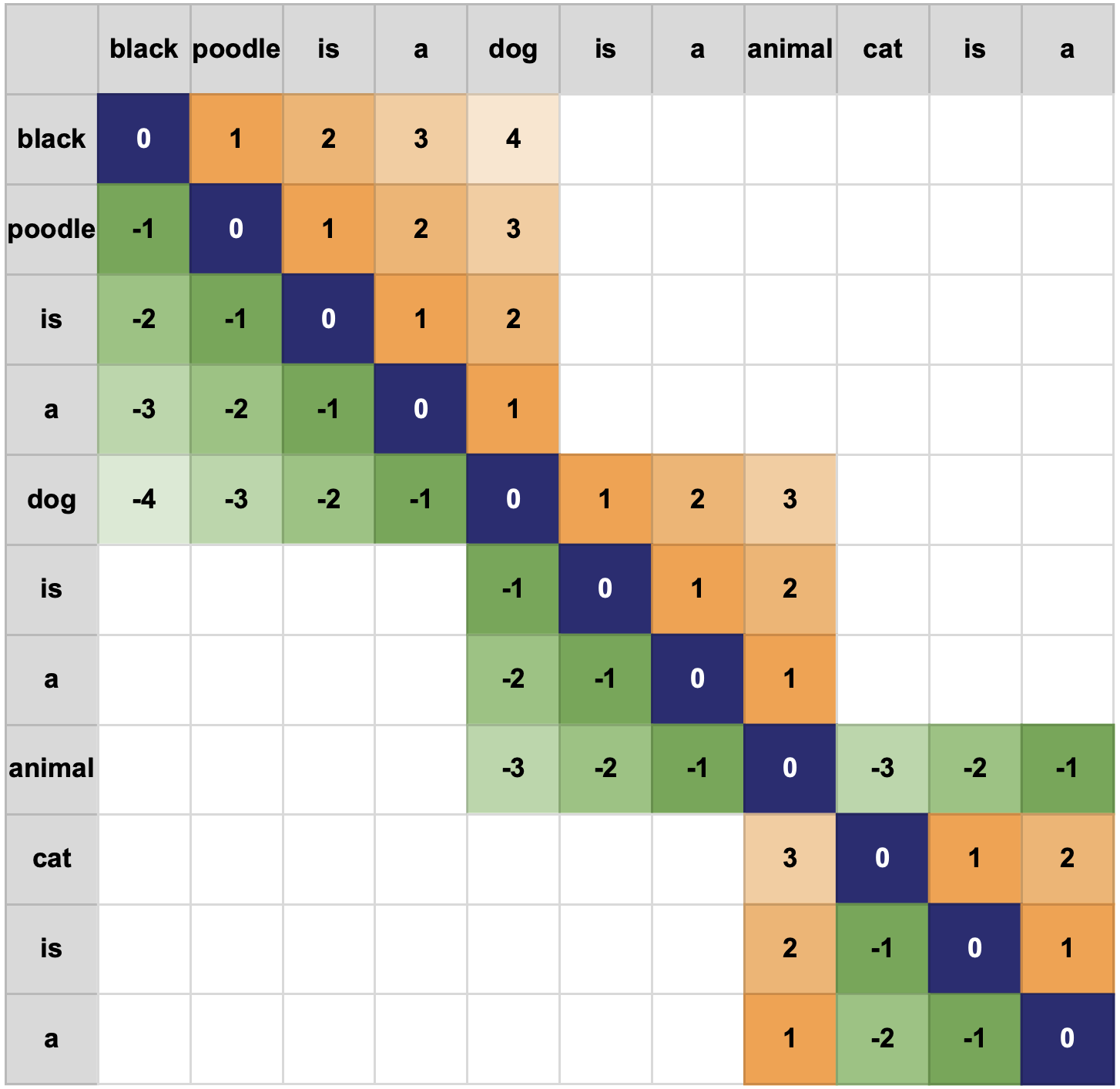}
         \caption{Relative position matrix $P$ for \lglm}
         \label{fig:app:lglm_P}
     \end{subfigure}
     \hspace{1.5cm}
     \centering
    \begin{subfigure}[b]{0.315\textwidth}
         \centering
         \includegraphics[width=\textwidth]{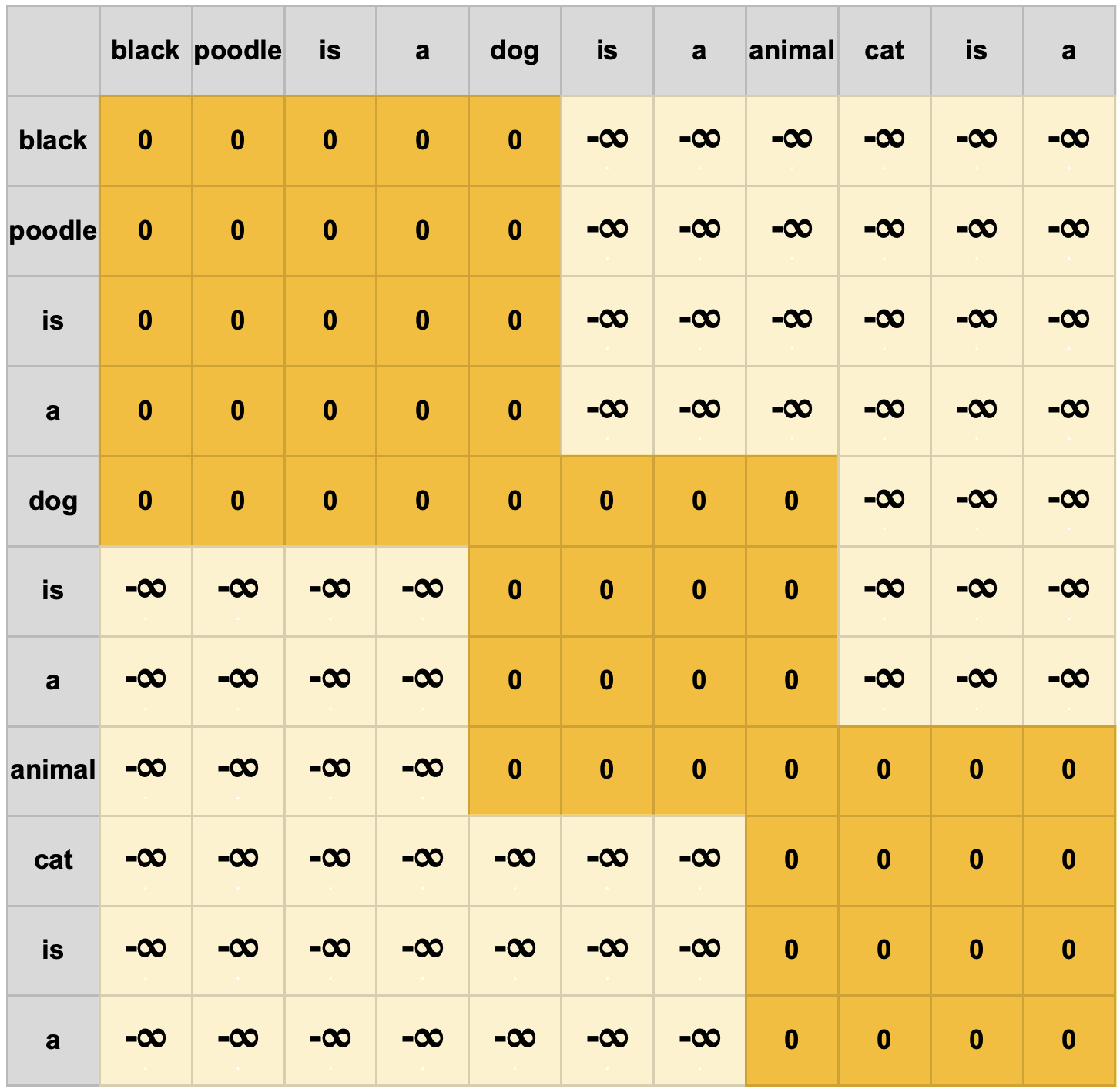}
         \caption{Mask matrix $M$ for \lglm.}
         \label{fig:app:lglm_M}
     \end{subfigure}
    \\ \vspace{0.4cm}
    \begin{subfigure}[b]{0.6\textwidth}
         \centering
         \includegraphics[width=\textwidth]{tikz/output-figure2} %
         \caption{Relative position $P$ for \textit{dog} in \gglm.}
         \label{fig:app:gglm_levi}
     \end{subfigure}
    \\ \vspace{0.4cm}
    \centering
    \begin{subfigure}[b]{0.315\textwidth}
         \centering
         \includegraphics[width=\textwidth]{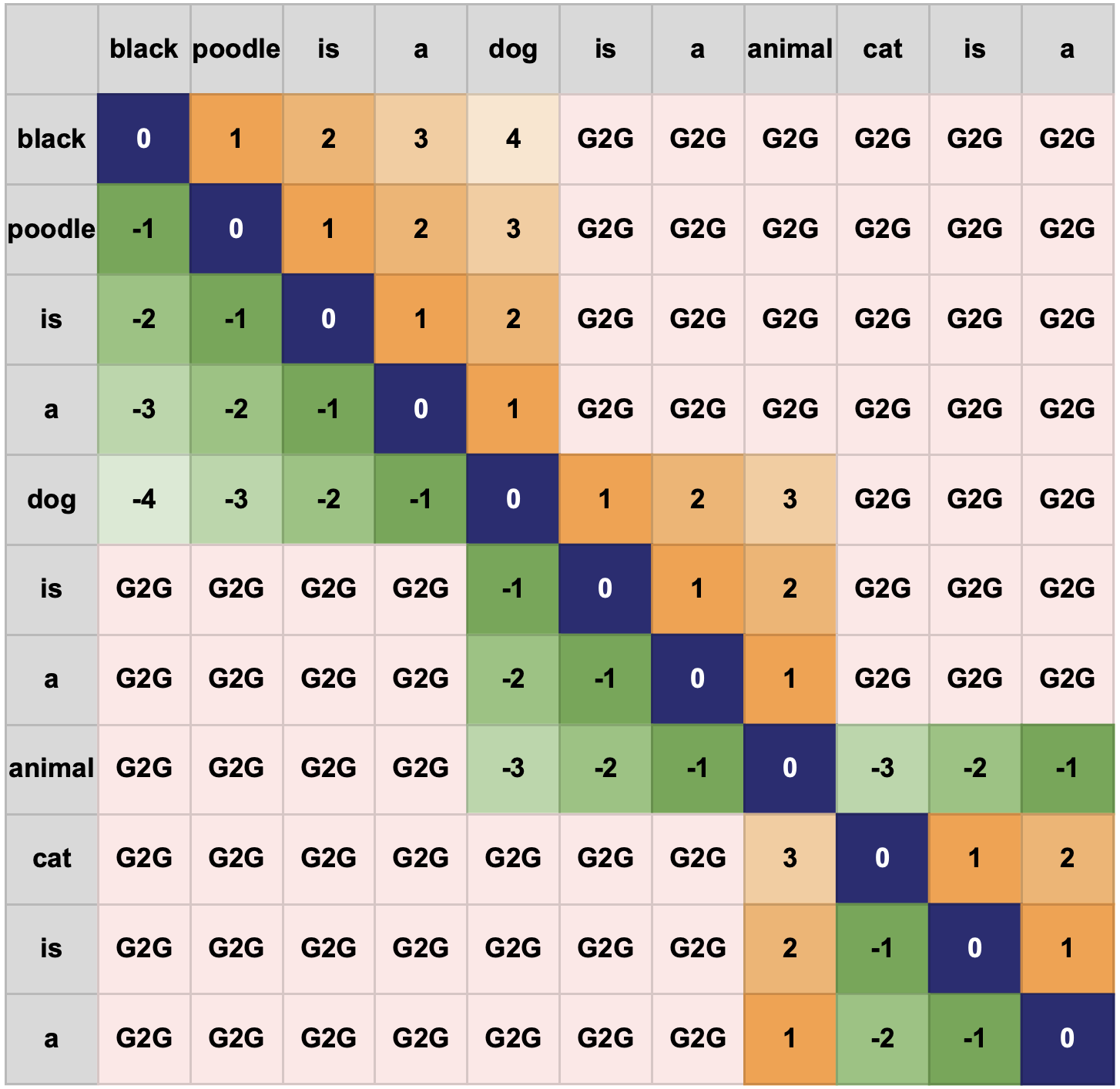}
         \caption{Relative position matrix $P$ for \gglm}
         \label{fig:app:gglm_P}
     \end{subfigure}
     \hspace{1.5cm}
     \centering
    \begin{subfigure}[b]{0.315\textwidth}
         \centering
         \includegraphics[width=\textwidth]{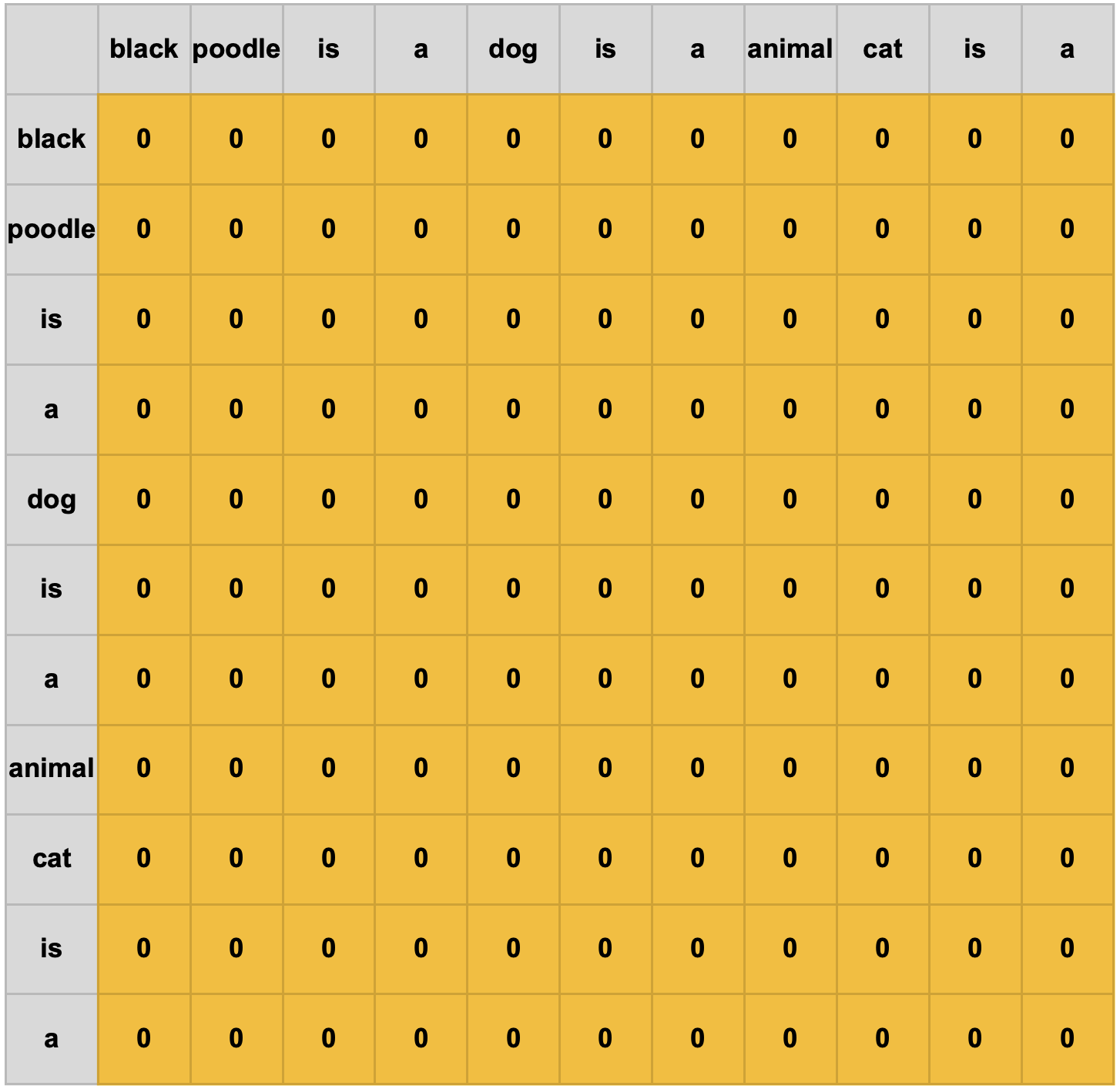}
         \caption{Mask matrix $M$ for \gglm.}
         \label{fig:app:gglm_M}
     \end{subfigure}
    \caption{Relative positions $P$ and masking $M$ for \lglm\ and \gglm.}
    \label{fig:app:PE_M}
\end{figure*}

\begin{figure*}  %
    \centering
    \begin{subfigure}[b]{0.45\textwidth}
         \centering
        \includegraphics[width=\textwidth]{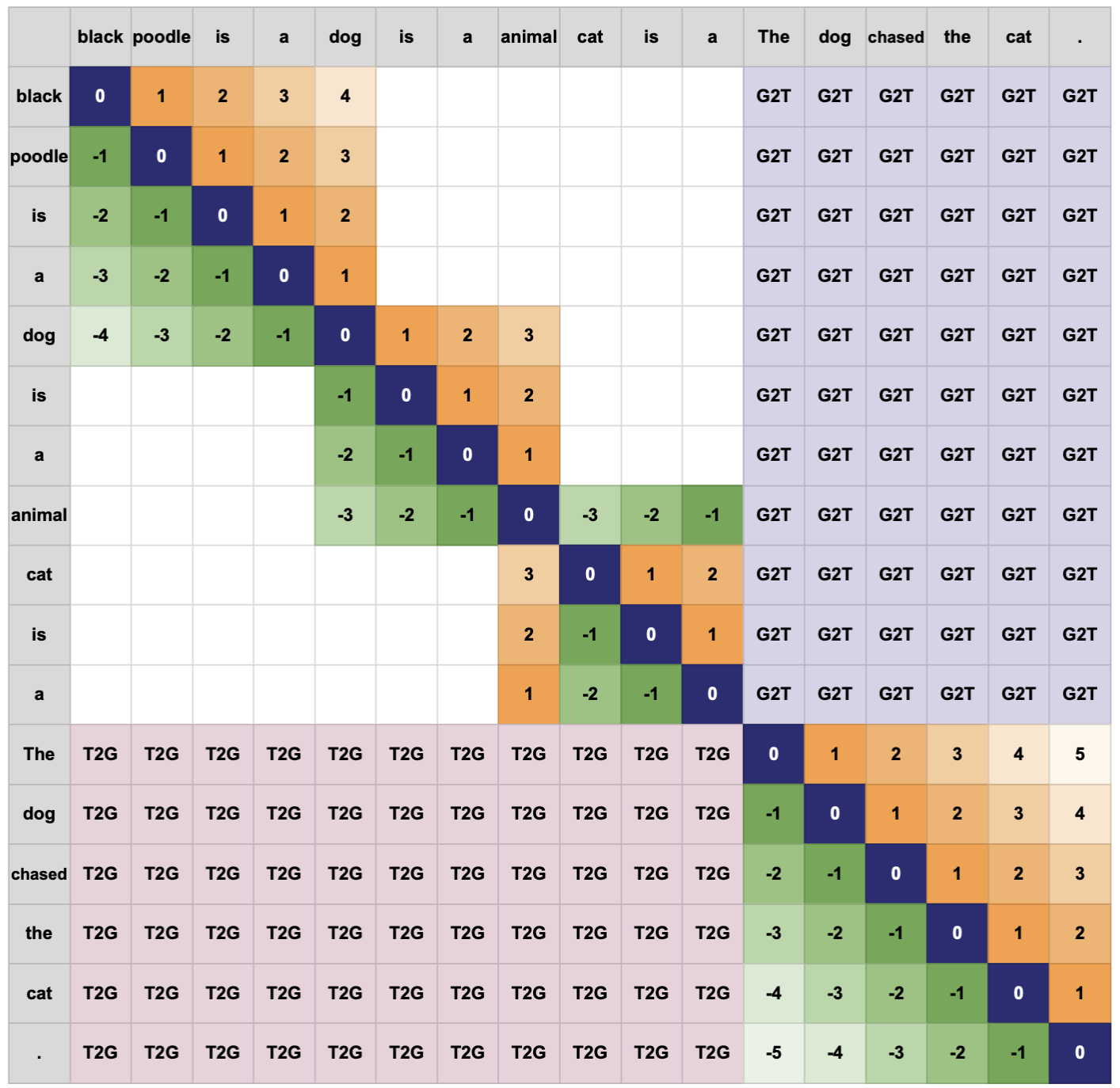}   
        \caption{Relative position matrix $P$ for \lglm.}
     \end{subfigure}
    \hfill
     \begin{subfigure}[b]{0.45\textwidth}
         \centering
        \includegraphics[width=\textwidth]{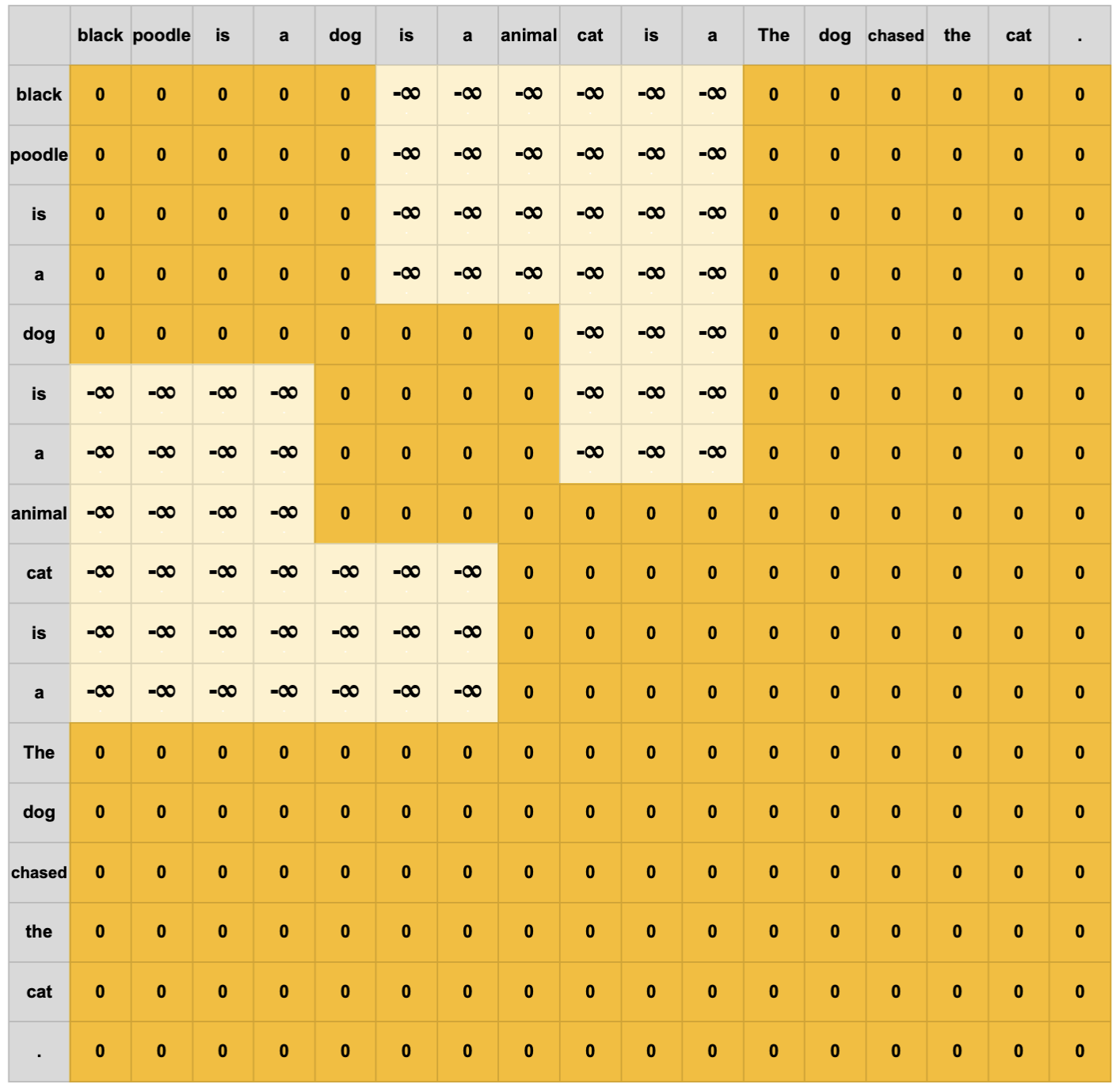}   
        \caption{Mask matrix $M$ for \lglm.}
     \end{subfigure}
    \\ \vspace{1cm}
     \begin{subfigure}[b]{0.45\textwidth}
         \centering
        \includegraphics[width=\textwidth]{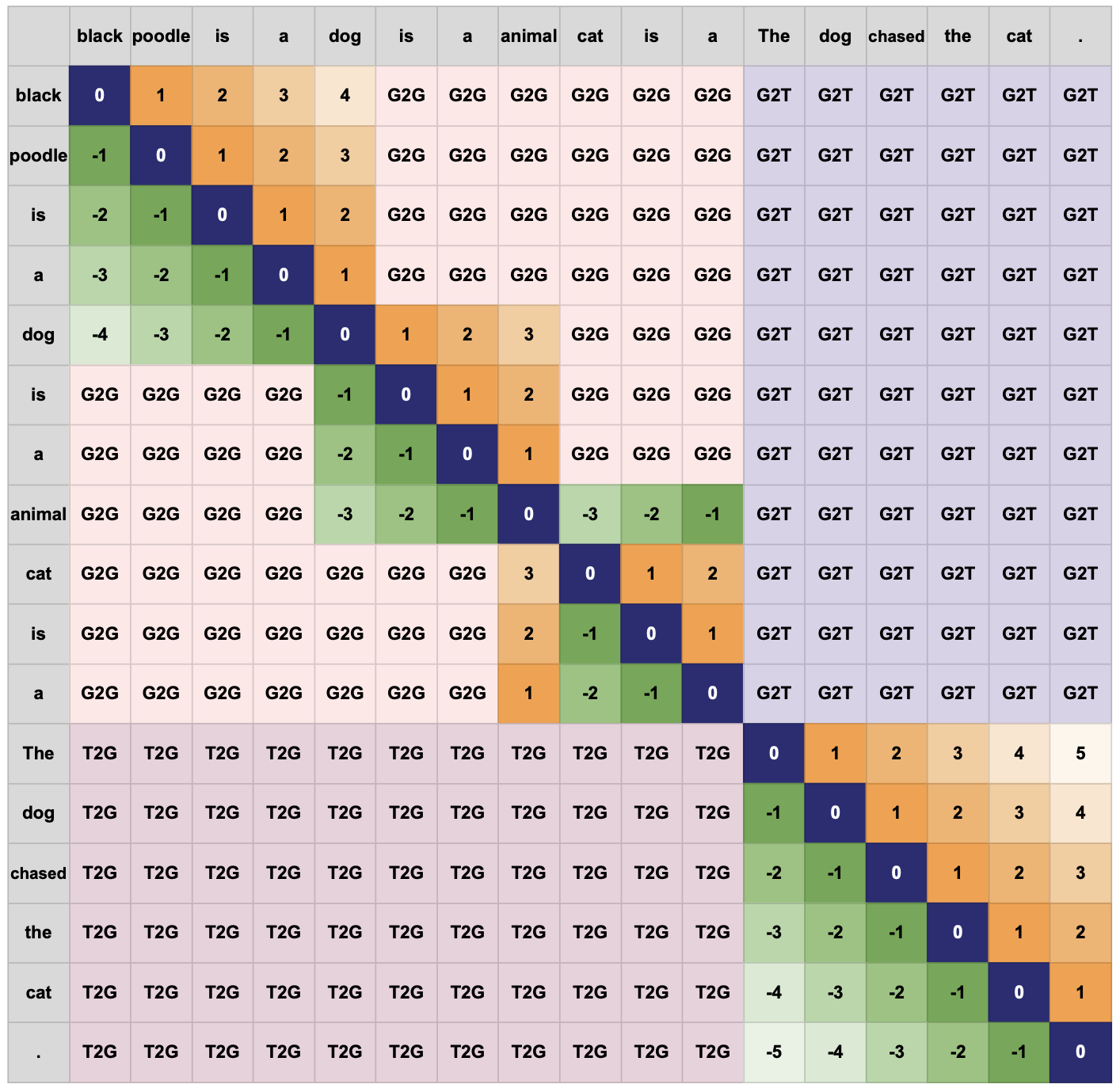}   
        \caption{Relative position matrix $P$ for \gglm.}
     \end{subfigure}
     \hfill
    \begin{subfigure}[b]{0.45\textwidth}
         \centering
        \includegraphics[width=\textwidth]{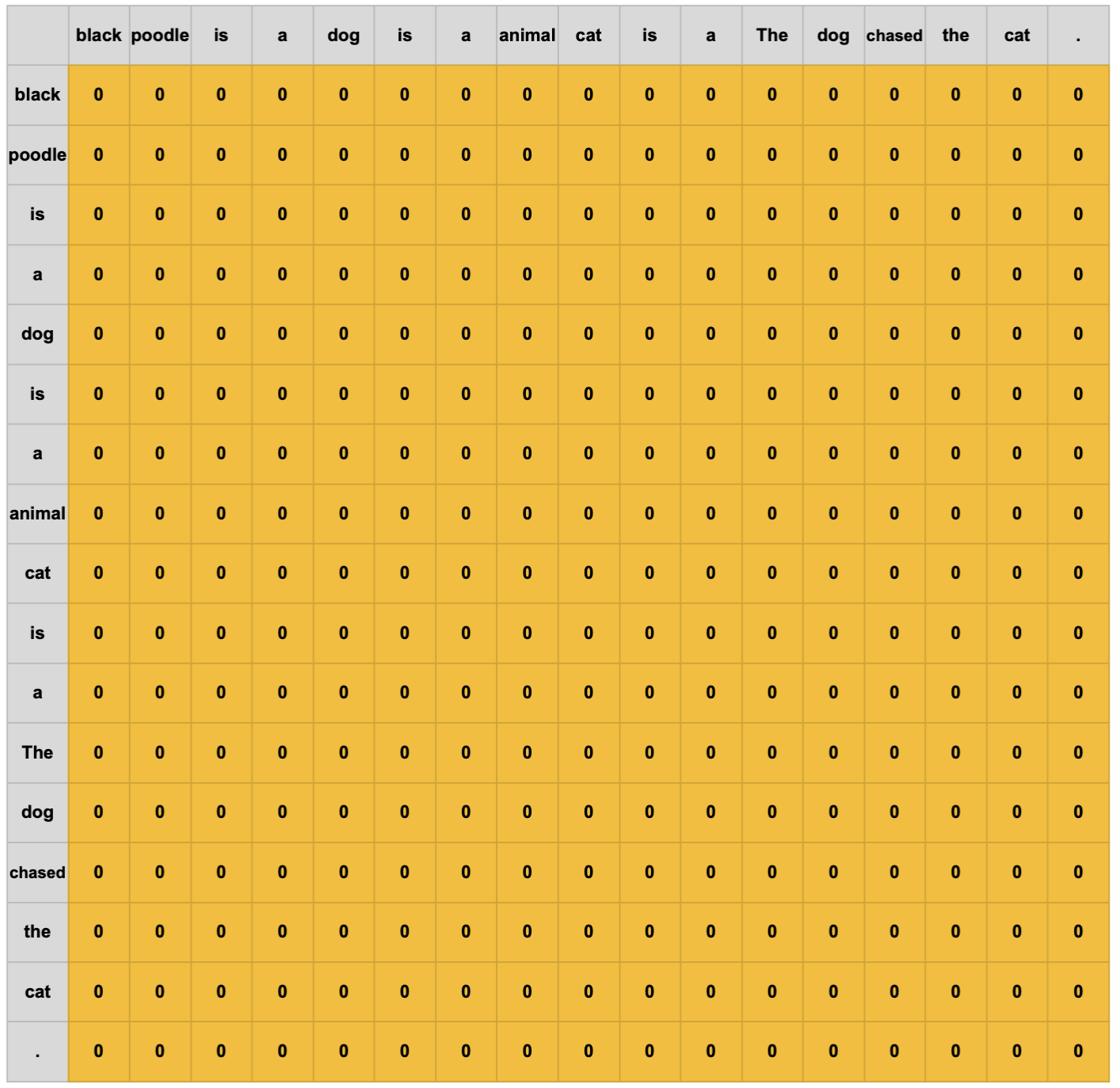}   
        \caption{Mask matrix $M$ for \gglm.}
     \end{subfigure}

    \caption{Relative positions $P$ and masking $M$ for \lglm\ and \gglm\ when encoding text and graph jointly. The example sentence is ``\textit{The dog chased the cat.}''}
    \label{fig:app:joint_PE_M}
\end{figure*}

Fig.~\ref{fig:app:PE_M} shows the matrices $P$ and $M$ for \lglm\ and \gglm. Fig.~\ref{fig:app:joint_PE_M} shows the same matrices for joint encoding of text and graph data. 

\section{Experiments} \label{sec:app:experiments}
\subsection{ConceptNet}
\subsubsection{Dataset} \label{sec:app:exp:cn:dataset} 
We experiment on randomly selected subgraphs from the largest connected component of the English part of CN version 5.7~\citep{speer-etal-2017-conceptnet}, which consists of 125,661 concepts and 1,025,802 triplets. We select 17 distinct relation label classes (cf. Tab.~\ref{tab:app:verbalization_templates_cn}), ensuring sufficient frequency and semantic dissimilarity. For each class, we randomly sample 1,000 triplets, allowing only cases where exactly one triplet connects the head and tail entities, to reduce label ambiguity. These 1,000 instances are split into train (800), dev (100), and test (100). This creates a balanced dataset of 13,600 train, 1,700 dev, and 1,700 test instances. To predict relation labels, we replace them with \texttt{<extra\_id\_0>}, T5's first mask token. For our experiments, we replace CN (unmasked) relations with more natural verbalizations. Tab.~\ref{tab:app:verbalization_templates_cn} shows the static verbalization for each relation. 

During graph construction we control the graph size, parameterized by the radius $r$. We start with a radius of $r=1$, when we consider only the two concepts (head and tail) in the target triplet. To create a larger graph context, we randomly select 4 adjacent triplets -- 2 for the head, and 2 for the tail entity of the original triplet. A graph with radius $r=2$ is formed by the subgraph spanned by all entities used in these 5 triplets. For $r=3$ we again randomly select 2 triplets for each of the outer (up to) 4 entities, yielding (up to) 13 triplets. To avoid accidentally adding more short-ranged information, we restrict the new triplets to triplets that actually extend the radius of the graph. This enables us to control graph size and complexity, while still enabling sufficient diversity in the graph structure. 
Further, the graphs are created such that graphs for smaller radii are strict subgraphs of graphs with larger radii. This ensures that performance changes with increasing radii are due to long-ranged connections, and not due to potentially different short-ranged information. Tab.~\ref{tab:app:datasetstatistics_relation} shows structural properties of CN subgraphs, depending on their radius. 

When masking subgraphs, we mask complete subgraphs of a certain size around the target to be predicted. The size of the masked subgraph is denoted by $m$, where $m=0$ means no masking, $m=1$ masks neighboring concepts, $m=2$ masks neighboring concepts and the next relations, and so on. Formally, $m$ denotes the radius of the masked graph in Levi representation, which should not be confused with the extended Levi graph, nor the normal graph representation. We replace each concept and relation in the masked subgraph with a different mask token. This in principle enables LM baselines to internally reconstruct  the graph. 

\begin{table*} %
    \centering
    \resizebox{.75\linewidth}{!}{
    \begin{tabular}{ccccccccccc}
    \toprule
        Metric & $r=1$ & $r=2$ & $r=3$ & $r=4$ & $r=5$ \\ \cmidrule(r){1-1} \cmidrule(l){2-6}
        \#nodes & 2.00 $\pm$ 0.00 & 5.77 $\pm$ 0.46 & 12.28 $\pm$ 1.67 & 23.47 $\pm$ 4.33 & 42.90 $\pm$ 9.57 \\
        \#edges & 1.00 $\pm$ 0.00 & 8.25 $\pm$ 2.74 & 19.19 $\pm$ 5.33 & 36.41 $\pm$ 9.09 & 66.06 $\pm$ 16.77 \\
        mean degree & 1.00 $\pm$ 0.00 & 2.87 $\pm$ 0.96 & 3.14 $\pm$ 0.82 & 3.11 $\pm$ 0.59 & 3.08 $\pm$ 0.42 \\
    \bottomrule
    \end{tabular}
    }
    \caption{Structural statistics of ConceptNet (\S\ref{sec:exp:cn}) train graphs.}
    \label{tab:app:datasetstatistics_relation}
\end{table*}

\begin{table}
    \centering
    \resizebox{.93\linewidth}{!}{
    \begin{tabular}{ccc}
    \toprule
    & Relation & Verbalization \\
    \cmidrule(r){1-2} \cmidrule(l){3-3}
    \parbox[t]{2mm}{\multirow{17}{*}{\rotatebox[origin=c]{90}{Used as relation label}}}
    & Antonym & is an antonym of \\
    & AtLocation & is in \\
    & CapableOf & is capable of \\
    & Causes & causes \\
    & CausesDesire & causes desire \\
    & DistinctFrom & is distinct from \\
    & FormOf & is a form of \\
    & HasContext & has context  \\
    & HasPrerequisite & has prerequisite  \\
    & HasProperty & is \\
    & HasSubevent & has subevent \\ 
    & IsA & is a \\
    & MannerOf & is a manner of \\
    & MotivatedByGoal & is motivated by \\
    & PartOf & is a part of \\
    & Synonym & is a synonym of \\
    & UsedFor & is used for \\ 
    \cmidrule(r){1-2} \cmidrule(l){3-3}
    \parbox[t]{2mm}{\multirow{16}{*}{\rotatebox[origin=c]{90}{Not used as relation label}}}
    & CreatedBy & is created by  \\
    & DefinedAs & is defined as \\
    & Desires & desires \\
    & Entails & entails  \\
    & HasA & has \\
    & HasFirstSubevent & starts with\\
    & HasLastSubevent & ends with \\
    & InstanceOf & is an instance of \\
    & LocatedNear & is near  \\
    & MadeOf & is made of  \\
    & NotCapableOf & is not capable of\\
    & NotDesires & does not desire \\
    & NotHasProperty & is not \\
    & ReceivesAction & receives action \\
    & RelatedTo & is related to\\
    & SymbolOf & is a symbol of \\
    \bottomrule
    \end{tabular}}
    \caption{Verbalization templates for relations in ConceptNet. The upper part of the relations are the 17 classes in the classification task.}
    \label{tab:app:verbalization_templates_cn}
\end{table}

\subsubsection{Experimental setup and baselines} \label{sec:app:exp:cn:setup}
Tab.~\ref{tab:app:hyperparameter} shows our hyperparameters. For the GNNs, we tested different numbers of layers (2, 3, 4, 5), hidden channel dimensions (32, 64, 128), and non-linearities (ReLU, leaky ReLU) in preliminary experiments.

\begin{table}
    \centering
    \resizebox{\linewidth}{!}{
    \begin{tabular}{clc}
        \toprule
        & Parameter & Value \\ 
        \cmidrule(r){1-2} \cmidrule(l){3-3}

        \parbox[t]{2mm}{\multirow{10}{*}{\rotatebox[origin=c]{90}{GLM, LM \& GT}}}
        & Loss & cross entropy loss \\
        & Optimizer & AdamW \\
        & Learning rate & \SI{1e-4}{} (FT) \& \SI{5e-3}{} (LP) \\
        & Batchsize & 32 \\
        & Max. \# epochs & 50 \\
        & Early stopping criterion & dev loss \\
        & Early stopping \# epochs & 5 \\
        & \# parameters in small & \phantom{0}35B (FT) \& \phantom{0}8k (LP)\phantom{0}\\
        & \# parameters in base & 110B (FT) \& 13k (LP)\phantom{0} \\
        & \# parameters in large & 335B (FT) \& 17k (LP)\phantom{0} \\
        & \# encoder layers in small & 6 \\
        & \# encoder layers in base & 12 \\
        & \# encoder layers in large & 24 \\
        \cmidrule(r){1-2} \cmidrule(l){3-3}
        \parbox[t]{2mm}{\multirow{10}{*}{\rotatebox[origin=c]{90}{GNN}}}
        & Loss & cross entropy loss \\
        & Optimizer & AdamW \\
        & Learning rate & \SI{5e-3}{} \\
        & Batchsize & 32 \\
        & Max. \# epochs & 50 \\
        & Early stopping criterion & dev loss \\
        & Early stopping \# epochs & 5 \\
        & \# layers & 3 \\
        & hidden channel dimension & 64 \\
        & non-linearity & ReLU \\
        \bottomrule
    \end{tabular}
    }
    \caption{Hyperparameters for \S\ref{sec:exp:cn}. FT stands for finetuning and LP stand for linear probing. ``\# parameters'' is the number of trainable parameters.}
    \label{tab:app:hyperparameter}
\end{table}

\subsubsection{Results} \label{sec:app:exp:cn:results}
Tab.~\ref{tab:res_rel_allparams_modelsize} shows performance on CN for different modelsizes. 
\begin{table*} %
    \centering
    \resizebox{\linewidth}{!}{
    \begin{tabular}{cccccccccccc}
    \toprule
        \multirow{2}{*}{Model} 
        & $r$ & 1 & 2 & 3 & 4 & 5 & 4 & 4 & 4 & 4 & 4 \\
        & $m$ & 0 & 0 & 0 & 0 & 0 & 1 & 2 & 3 & 4 & 5 \\ \cmidrule(r){1-2} \cmidrule(lr){3-7} \cmidrule(l){8-12}
        
        \multirow{3}{*}{\lglm}
        & small & %
        64.0$\pm$1.3 & 64.0$\pm$1.0 & 64.4$\pm$0.7 & 64.1$\pm$0.9 & 64.2$\pm$1.1 & 47.9$\pm$0.4 & 26.8$\pm$0.8 & 23.8$\pm$0.9 & 19.8$\pm$1.1 & 18.1$\pm$0.7 \\
        & base & %
        67.6$\pm$0.8 & 69.6$\pm$0.9 & 69.8$\pm$0.5 & 69.8$\pm$1.3 & 69.6$\pm$0.7 & \textbf{49.2$\pm$0.8} & 29.3$\pm$0.8 & 24.4$\pm$0.3 & \textbf{20.8$\pm$0.9} & 19.6$\pm$0.8 \\
        & large & %
        \textbf{72.0$\pm$1.0} & \textbf{71.4$\pm$1.5} & \textbf{72.2$\pm$1.0} & \textbf{72.7$\pm$0.8} & \textbf{71.5$\pm$1.8} & 48.4$\pm$1.1 & \textbf{29.7$\pm$1.6} & \textbf{24.8$\pm$1.6} & 20.0$\pm$0.9 & \highlight{\textbf{20.3$\pm$0.5}} \\ 
        \cmidrule(r){1-2} \cmidrule(lr){3-7} \cmidrule(l){8-12}
        
        \multirow{3}{*}{\gglm}
        & small & %
        63.2$\pm$0.9 & 64.4$\pm$1.1 & 64.6$\pm$1.2 & 64.1$\pm$1.3 & 65.3$\pm$0.7 & 48.0$\pm$0.6 & 27.2$\pm$0.7 & 24.2$\pm$0.7 & 20.2$\pm$1.4 & 19.2$\pm$0.7 \\
        & base & %
        67.8$\pm$0.7 & 71.3$\pm$1.0 & 70.5$\pm$1.2 & 71.5$\pm$1.1 & 71.1$\pm$0.4 & 49.7$\pm$1.2 & 30.2$\pm$0.8 & \highlight{\textbf{25.5$\pm$0.8}} & \highlight{\textbf{21.4$\pm$1.2}} & \textbf{20.1$\pm$0.2} \\
        & large & %
        \textbf{72.1$\pm$1.1} & \highlight{\textbf{73.9$\pm$0.7}} & \highlight{\textbf{74.2$\pm$0.6}} & \highlight{\textbf{74.8$\pm$0.8}} & \highlight{\textbf{73.9$\pm$0.7}} & \highlight{\textbf{50.1$\pm$0.5}} & \highlight{\textbf{31.9$\pm$1.2}} & 24.4$\pm$1.5 & 21.2$\pm$0.6 & 19.6$\pm$0.8\\ 
        \cmidrule(r){1-2} \cmidrule(lr){3-7} \cmidrule(l){8-12}

        \multirow{3}{*}{T5 list}
        & small & %
        64.9$\pm$1.0 & 64.9$\pm$1.2 & 64.9$\pm$1.3 & 63.9$\pm$0.9 & 64.0$\pm$0.6 & 40.4$\pm$0.8 & 21.8$\pm$0.8 & 17.8$\pm$1.0 & 15.4$\pm$0.3 & 12.8$\pm$0.5 \\
        & base & %
        71.2$\pm$0.9 & 69.5$\pm$0.7 & 69.5$\pm$1.0 & 70.4$\pm$1.6 & 70.4$\pm$0.7 & 40.7$\pm$0.9 & 25.5$\pm$1.2 & 17.8$\pm$0.2 & 16.4$\pm$1.3 & \textbf{13.9$\pm$0.7} \\
        & large & %
        \textbf{74.5$\pm$0.4} & \textbf{73.7$\pm$0.4} & \textbf{73.5$\pm$0.6} & \textbf{73.6$\pm$0.8} & \textbf{73.3$\pm$1.0} & \textbf{41.2$\pm$1.5} & \textbf{27.9$\pm$1.0} & \textbf{18.3$\pm$0.9} & \textbf{17.0$\pm$0.5} & 13.0$\pm$0.9 \\ 
        \cmidrule(r){1-2} \cmidrule(lr){3-7} \cmidrule(l){8-12}

        \multirow{3}{*}{T5 set}
        & small & %
        63.9$\pm$0.7 & 65.8$\pm$0.8 & 64.0$\pm$0.3 & 64.1$\pm$1.2 & 64.3$\pm$1.1 & 40.3$\pm$1.2 & 21.8$\pm$0.7 & 18.0$\pm$0.6 & 15.5$\pm$0.6 & 13.1$\pm$0.7 \\
        & base & %
        71.2$\pm$0.6 & 69.8$\pm$0.6 & 69.5$\pm$0.6 & 70.1$\pm$0.7 & 69.8$\pm$1.4 & 40.4$\pm$0.9 & 23.9$\pm$1.1 & \textbf{18.5$\pm$1.1} & \textbf{16.3$\pm$0.3} & \textbf{14.3$\pm$0.7} \\
        & large & %
        \textbf{\highlight{74.9$\pm$0.3}} & \textbf{73.0$\pm$0.5} & \textbf{73.1$\pm$0.8} & \textbf{72.5$\pm$1.1} & \textbf{73.5$\pm$0.4} & \textbf{41.2$\pm$1.3} & \textbf{25.1$\pm$1.3} & 17.4$\pm$0.9 & 15.9$\pm$0.5 & 13.2$\pm$0.8\\
    \bottomrule
    \end{tabular}
    }
    \caption{Relation label classification accuracy on ConceptNet (\S\ref{sec:exp:cn}) when training all parameters. Best score per model family is boldfaced, and best score overall is highlighted in yellow.}
    \label{tab:res_rel_allparams_modelsize}
\end{table*}

\subsection{Wikidata and Wikipedia}
\subsubsection{Dataset} \label{sec:app:exp:wiki:dataset}
\citet{huguet-cabot-navigli-2021-rebel} propose a large-scale corpus of aligned Wikipedia abstracts and Wikidata~\citep{vrandecic-etal-2014-wikidata} triplets. They first extract Wikidata entities from the abstract, and then link these entities with triplets in Wikidata. They are interested in triplets that are entailed by the text, so they use a NLI model to filter out all other triplets. They publicly released the extracted entities and the filtered triplets. 

For our purpose, we are interested in aligned graphs and texts, but triplets in the graph do not necessarily have to be entailed by the text. Hence, we find all triplets between the extracted entities using the Wikidata Query Service.\footnote{\url{https://query.wikidata.org}, accessed in Jan. 2024.} From \citet{huguet-cabot-navigli-2021-rebel} we know which triplets in our graphs are entailed by the text. %

Similar to \citet{huguet-cabot-navigli-2021-rebel} we consider the 220 most common relations in the train split as our relation labels. Additionally, we add a ``no-relation'' label, yielding 221 relation classes. 

For \SI{10}{\%} of the graphs we randomly add a new triplet between previously unconnected head and tail entity, and the mask token as relation. For these graphs ``no-relation'' is the correct relation label. For the other \SI{90}{\%} graphs we replace a random existing relation with the mask token, while making sure that (i) the existing relation is in our 220 labels and that (ii) there is no other triplet connecting the respective head and tail entities. We remove instances where no suitable triplet is available. This yields a dataset with 2,449,582 train, 135,828 val and 135,923 test instances. 

\begin{figure*}
    \centering
    \begin{subfigure}[b]{0.49\textwidth}
         \centering
         \includegraphics[width=\textwidth]{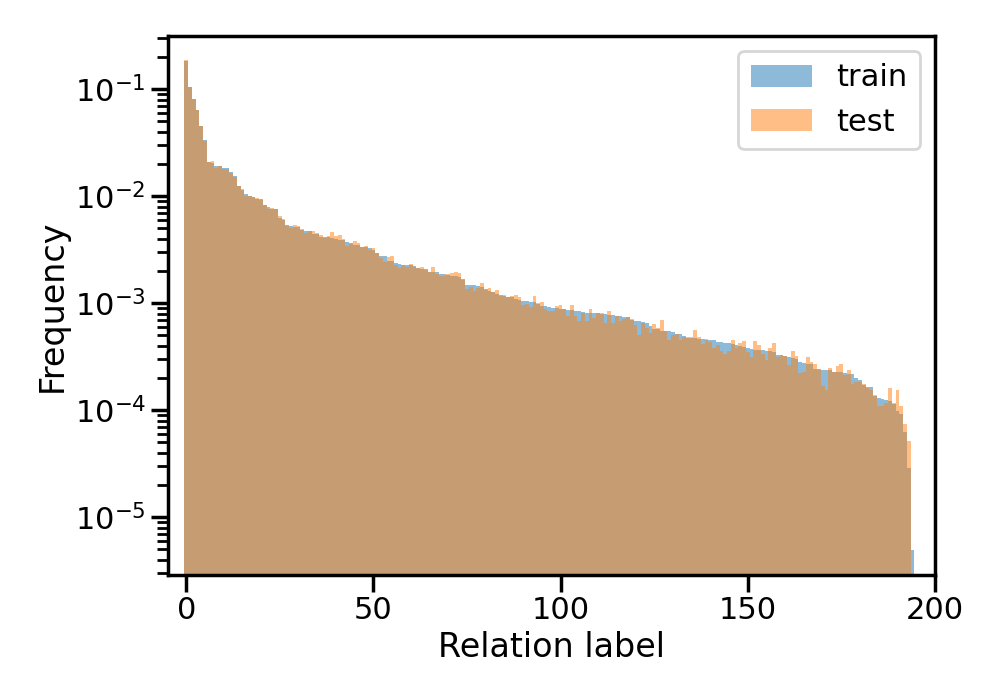}
         \caption{Relation label.}
     \end{subfigure}
     \hfill
     \begin{subfigure}[b]{0.49\textwidth}
         \centering
         \includegraphics[width=\textwidth]{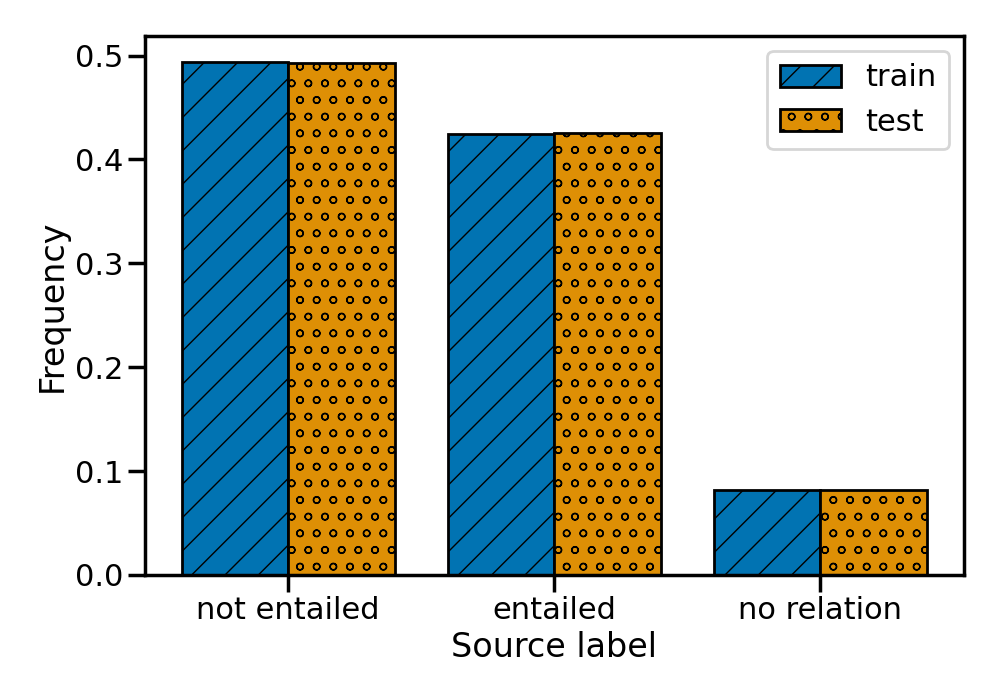}
         \caption{Source label.}
     \end{subfigure}

    \caption{Label distributions for Wikidata (\S\ref{sec:exp:wiki}) train and test sets.}
    \label{fig:app:exp:wiki:distributions}
\end{figure*}

Fig.~\ref{fig:app:exp:wiki:distributions} shows the label distributions for relation and source for train and test. Out of the 221 relations, only 195 and 194 relations occur in the train and test set, respectively. All relations in the test set also occur in the train set. 

Tab.~\ref{tab:app:exp:wikidata:datasetstatistics_structure} shows graph statistics. Compared to CN subgraphs (c.f.~Tab.~\ref{tab:app:datasetstatistics_relation}) the graphs are relatively small, matching the size of $r=2$. On CN we found that LMs can perform well on such small graphs, so we expect that the performance gap between GLMs and LM baselines on Wikidata would be larger if Wikidata subgraphs were larger. 

\begin{table} %
    \centering
    \resizebox{.9\linewidth}{!}{
    \begin{tabular}{ccc}
    \toprule
        Metric & train & test \\ \cmidrule(r){1-1} \cmidrule(l){2-3}
        \#nodes & 5.59 $\pm$ \phantom{0}3.77 & 5.60 $\pm$ \phantom{0}3.78 \\
        \#edges & 8.71 $\pm$ 11.99 & 8.71 $\pm$ 12.01 \\
        mean degree & 2.66 $\pm$ \phantom{0}1.58 & 2.66 $\pm$ \phantom{0}1.58 \\
    \bottomrule
    \end{tabular}
    }
    \caption{Structural statistics of Wikidata (\S\ref{sec:exp:wiki}) subgraphs.}
    \label{tab:app:exp:wikidata:datasetstatistics_structure}
\end{table}

\subsubsection{Experimental setup and baselines} \label{sec:app:exp:wiki:setup}
For these experiments we omit GNNs as a baseline, since they can't natively process texts. 

The other models all compute an embedding of the mask token, and then two separate classification heads produce predictions for the relations (221 classes) and the source (3 classes). For each prediction, we compute the Cross Entropy Loss. The final loss is the weighted sum of these losses, weighted by 0.9 and 0.1 respectively. The relation classification has a higher weight since it has many more classes and hence, is potentially more difficult. This means that model parameters are optimized for both objectives jointly, while only the linear classification heads can specialize on their respective task. 

The dataset is unbalanced (c.f.\ Fig~\ref{fig:app:exp:wiki:distributions}), so report macro F1 scores instead of accuracy. This means that models only achieve high scores if they perform well on all classes, including minority classes. 

We assume that classifying one out of 221 relations requires fine grained text understanding, so we initialize models from T5-large instead of T5-small. To reduce computational load, we only train one model per setting. Further, we enable efficient batching by restricting inputs to a maximum of 512 tokens. This truncates \SI{2.8}{\%} of train instances for GLMs and \SI{5.1}{\%} for LM baselines due to their less efficient graph encoding. 

Hyperparameters are identical to Tab.~\ref{tab:app:hyperparameter}, except that (i) we reduce batch size to 8, (ii) train for at most 1 epoch and (iii) don't use early stopping. 

\subsubsection{Results} \label{sec:app:exp:wiki:res}
\begin{figure*}
    \centering

    \begin{subfigure}[b]{0.45\textwidth}
         \centering
         \includegraphics[width=\textwidth]{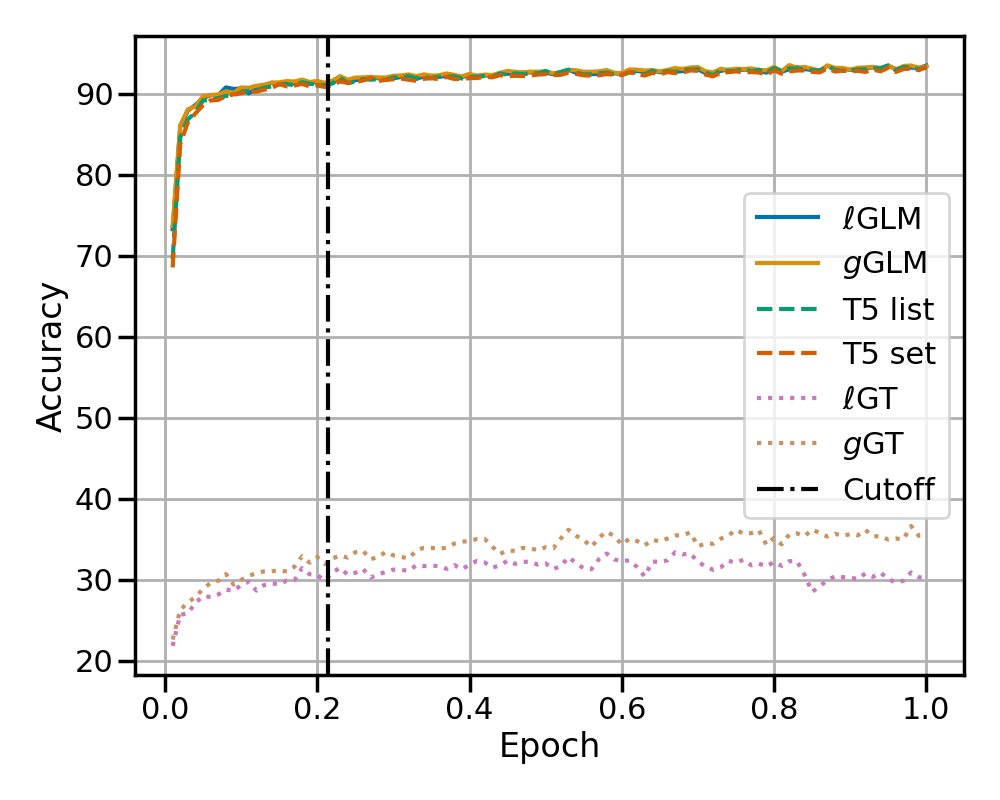}
         \caption{Evaluation on train set.}
         \label{fig:app:exp:wiki:long_train_curves_train}
     \end{subfigure}
     \hfill
     \begin{subfigure}[b]{0.45\textwidth}
         \centering
         \includegraphics[width=\textwidth]{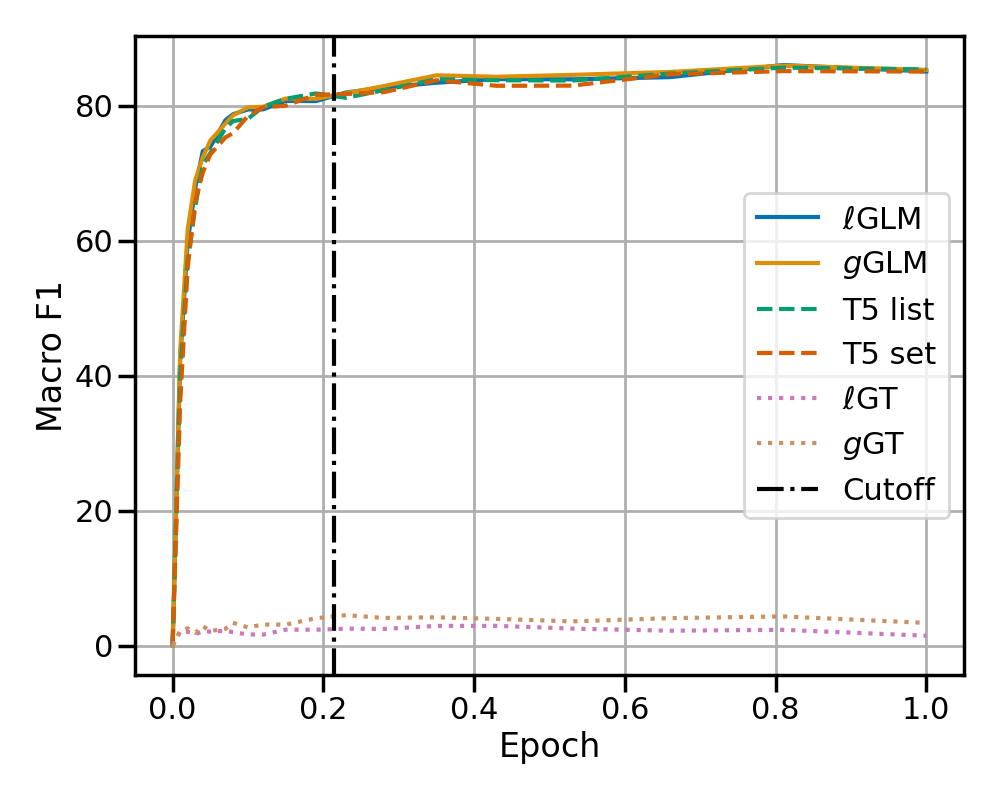}
         \caption{Evaluation on test set.}
     \end{subfigure}

    \caption{Training curves (\S\ref{sec:exp:wiki}) when training for a whole epoch, i.e., 2,449,582 train instances. Performances are for relation classification. On the train set we did not compute macro F1, so we report accuracy instead.}
    \label{fig:app:exp:wiki:long_train_curves}
\end{figure*}

Fig.~\ref{fig:app:exp:wiki:long_train_curves} shows the training curve when training for an entire epoch, i.e., 2,499,582. We observe that performances plateau beyond $\sim$ 0.2 epochs, so we stop training after 524,288 instances in our other experiments. 

Tab.~\ref{tab:app:exp:wiki:scores} shows concrete numbers for the models in Figures~\ref{fig:exp:wiki:baselines} and \ref{fig:app:exp:wiki:long_train_curves}. 

\begin{figure*}
    \centering

    \begin{subfigure}[b]{0.45\textwidth}
         \centering
         \includegraphics[width=\textwidth]{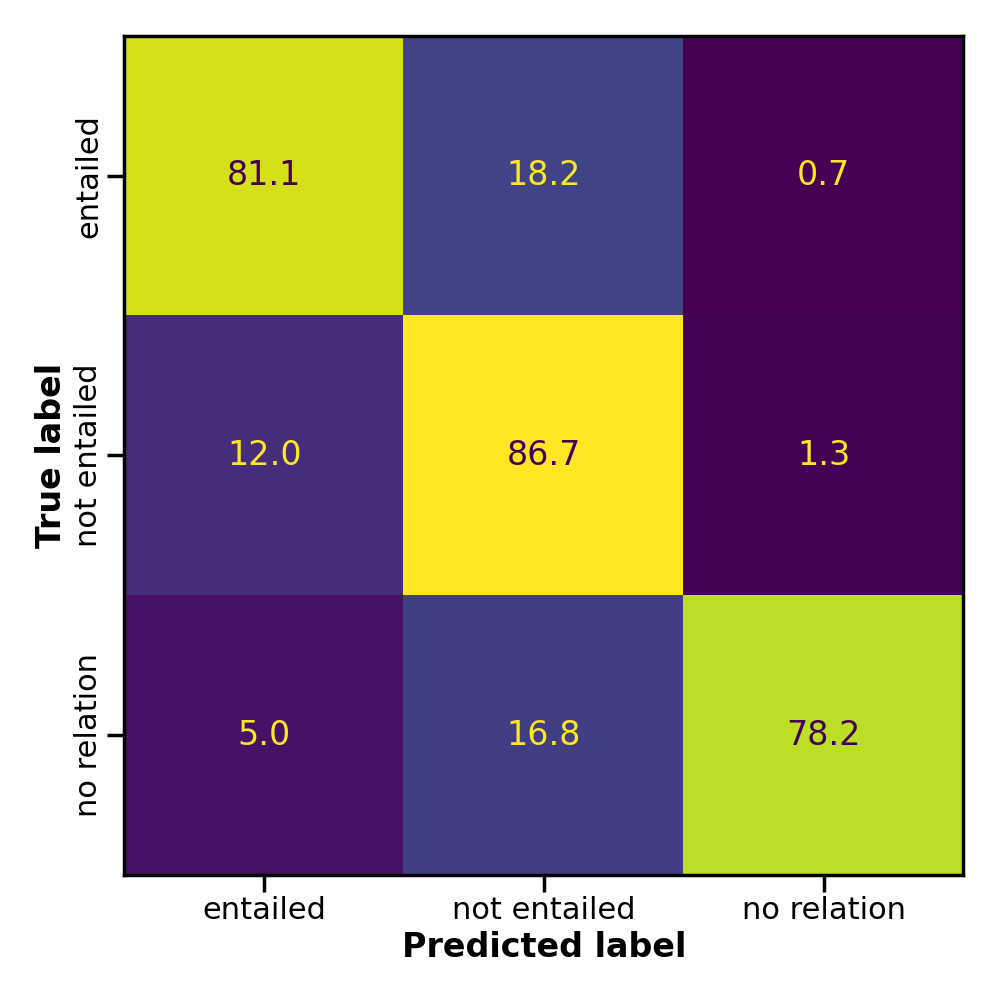}
         \caption{\lglm.}
     \end{subfigure}
     \hfill
     \begin{subfigure}[b]{0.45\textwidth}
         \centering
         \includegraphics[width=\textwidth]{img/wikidata-source_confusion_matrix-gGLM_524288_TTTF.png}
         \caption{\gglm.}
     \end{subfigure}

    \caption{Confusion matrices source prediction on Wikidata (\S\ref{sec:exp:wiki}).}
    \label{fig:app:exp:wiki:source_confusionmatrix}
\end{figure*}

Fig.~\ref{fig:app:exp:wiki:source_confusionmatrix} shows confusion matrices for source prediction. 

\begin{figure*}
    \centering

    \begin{subfigure}[b]{0.45\textwidth}
         \centering
         \includegraphics[width=\textwidth]{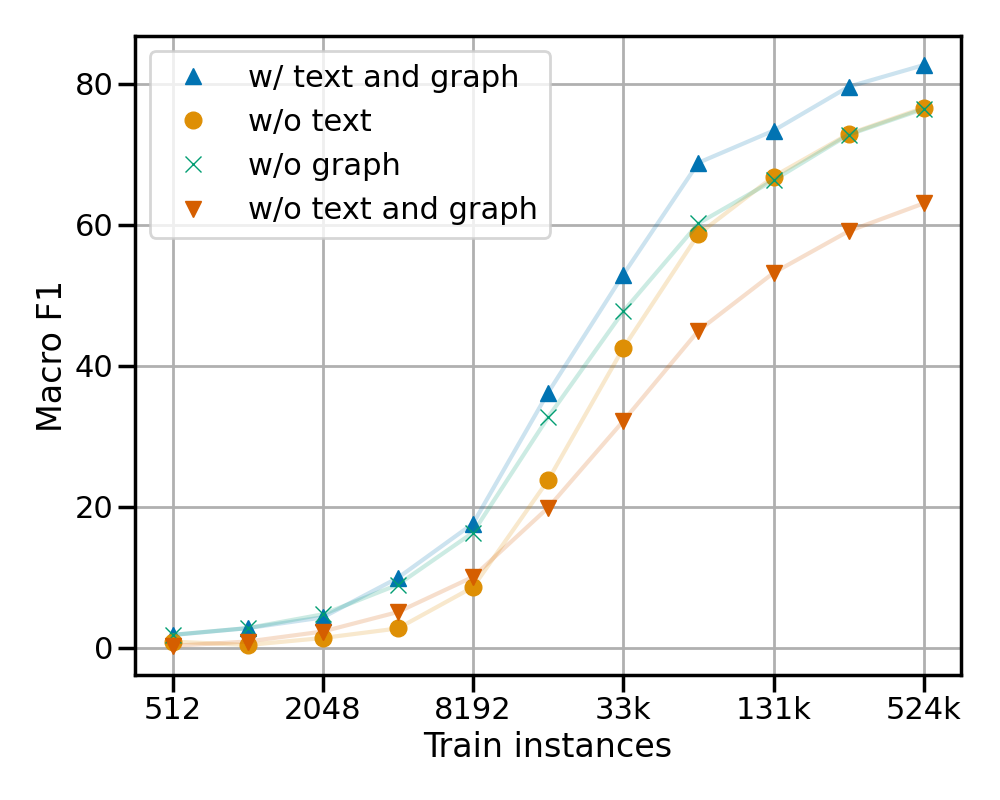}
         \caption{\lglm.}
     \end{subfigure}
     \hfill
     \begin{subfigure}[b]{0.45\textwidth}
         \centering
         \includegraphics[width=\textwidth]{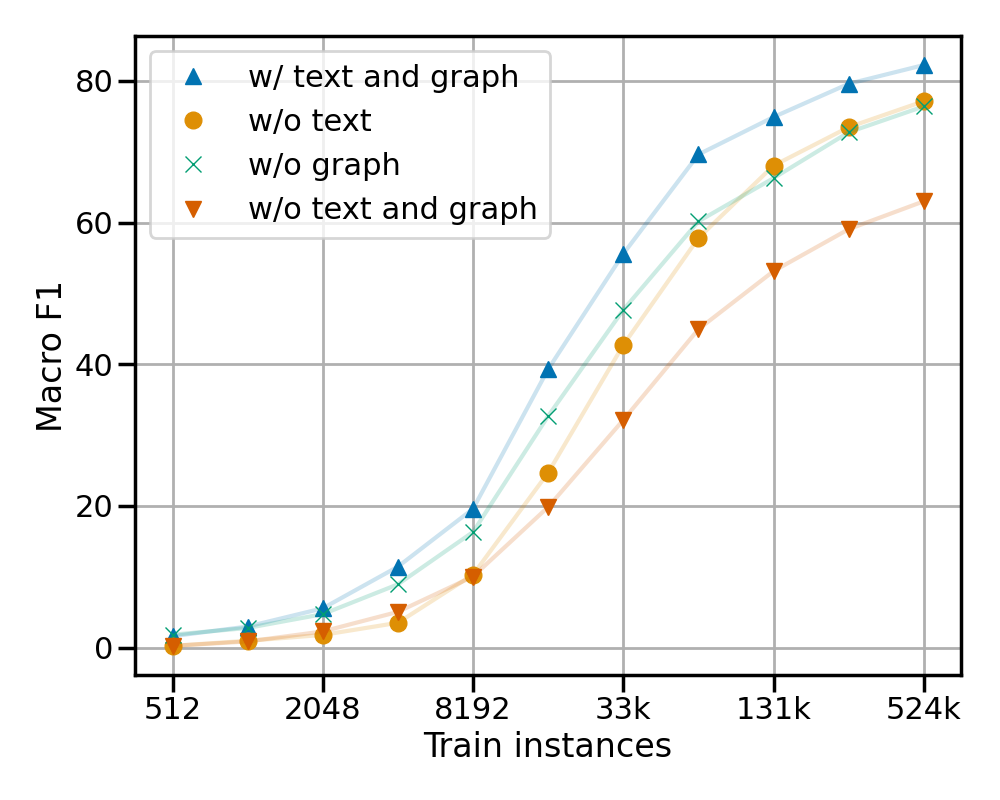}
         \caption{\gglm.}
     \end{subfigure}

    \caption{Ablation of different input modalities to GLMs. All runs are done without source prediction (besides \lglm\ and \gglm). Scores are for relation classification on Wikidata (\S\ref{sec:exp:wiki}).}
    \label{fig:app:exp:wiki:ablation}
\end{figure*}

Fig.~\ref{fig:app:exp:wiki:ablation} shows the test performance in relation classification of ablated models during different training steps. Table~\ref{tab:app:exp:wiki:ablation_text_graph} shows relation classification scores for (i) triplets entailed by text and for (ii) other triplets. 

\begin{table}
    \centering
    \resizebox{1.0\linewidth}{!}{
    \begin{tabular}{ccccc}
    \toprule
    \multirow{2}{*}{Model} & \multicolumn{2}{c}{524,288 train instances} & \multicolumn{2}{c}{2,449,582 train instances} \\
    & Relation & Source & Relation & Source \\
    \cmidrule(r){1-1} \cmidrule(lr){2-3} \cmidrule(lr){4-5}
    \lglm & 82.35 & 83.39 & 85.06 & 86.20 \\
    \gglm & 81.98 & 83.21 & 85.28 & 86.17 \\
    T5 list & 81.45 & 82.17 & 85.36 & 85.83 \\
    T5 set & 81.29 & 82.00 & 85.04 & 85.53 \\
    \lgt & \phantom{0}3.19 & 39.81 & \phantom{0}1.50 & 37.83 \\
    \ggt & \phantom{0}3.47 & 39.58 & \phantom{0}3.40 & 39.37 \\
        
    \bottomrule
    \end{tabular}
    }
    \caption{Macro F1 scores on Wikidata test set for relation classification and source classification. Scores are shown for models after training on different numbers of train instances.}
    \label{tab:app:exp:wiki:scores}
\end{table}

\begin{table} %
    \centering
    \resizebox{.9\linewidth}{!}{
    \begin{tabular}{lcccc}
    \toprule
        \multirow{2}{*}{Ablation} & \multicolumn{2}{c}{Entailed} & \multicolumn{2}{c}{Not entailed} \\
        & \lglm & \gglm & \lglm & \gglm \\
        \cmidrule(r){1-1} \cmidrule(lr){2-3} \cmidrule(l){4-5}
        w/ text \& graph & 85.46 & 84.85 & 78.47 & 78.46 \\
        w/o text & -8.40 & -6.75 & -4.57 & -4.28 \\
        w/o graph & -4.56 & -3.94 & -7.56 & -7.55 \\
        w/o text \& graph & -20.52\phantom{-} & -19.90\phantom{-} & -20.08\phantom{-} & -20.07\phantom{-} \\
    \bottomrule
    \end{tabular}
    }
    \caption{Ablations for KG population (\S\ref{sec:exp:wiki}). Scores are macro F1 for relation label classification on (i) triplets that are entailed by the text and (ii) all other triplets. Models are trained w/o source prediction.}
    \label{tab:app:exp:wiki:ablation_text_graph}
\end{table}

\section{Usage of AI assistants} 
We use GitHub Copilot (\url{https://github.com/features/copilot}) for speeding up programming, and ChatGPT 3.5 (\url{https://chat.openai.com}) to aid with reformulations. The content of this work is our own, and not inspired by AI assistants.

\end{document}